\documentclass[10pt,twocolumn,letterpaper]{article}
\usepackage{appendix}
\usepackage{iccv}
\usepackage{times}
\usepackage{epsfig}
\usepackage{graphicx}
\usepackage{amsmath}
\usepackage{amssymb}
\usepackage{color}
\usepackage{epsfig}
\usepackage{bm}
\usepackage{enumerate}
\usepackage{multirow}
\usepackage{amsthm}
\usepackage{array}
\usepackage{bigstrut}
\usepackage{algorithmic}
\usepackage{algorithm}

\usepackage[table]{xcolor}
\usepackage[font={small}]{caption}
\usepackage[caption=false,font=footnotesize]{subfig}
\usepackage[pagebackref=true,breaklinks=true,letterpaper=true,colorlinks,bookmarks=false]{hyperref}
\makeatletter
\newcommand\notsotiny{\@setfontsize\notsotiny{6}{7}}

\newcommand{\comment}[1]{}
\newcommand{\deflen}[2]{%      
    \expandafter\newlength\csname #1\endcsname
    \expandafter\setlength\csname #1\endcsname{#2}%
}
\definecolor{Gray}{gray}{0.9}
\newcolumntype{a}{>{\columncolor{Gray}}c}
\newcolumntype{b}{>{\columncolor{white}}c}
\newcommand{\printfnsymbol}[1]{%
  \textsuperscript{\@fnsymbol{#1}}%
}

\iccvfinalcopy % *** Uncomment this line for the final submission

\def\iccvPaperID{3514} % *** Enter the ICCV Paper ID here

\ificcvfinal\fi
\makeatother

\begin{document}
\deflen{inter}{1pt}
\deflen{widthdef}{55pt}
%%%%%%%%% TITLE
\title{Deep End-to-End Alignment and Refinement for Time-of-Flight RGB-D Module}

\author{Di Qiu\textsuperscript{1,2}\thanks{Both authors contributed equally. Jiahao Pang is the corresponding author, this work was done while he was with SenseTime.}
~~~~~~~
Jiahao Pang\textsuperscript{1}\printfnsymbol{1}
~~~~~~~
Wenxiu Sun\textsuperscript{1}
~~~~~~~
Chengxi Yang\textsuperscript{1}\\
\textsuperscript{1} SenseTime Research
~~~~~
\textsuperscript{2} The Chinese University of Hong Kong\\
{\small \tt sylvesterqiu@gmail.com, jpang@connect.ust.hk, \{sunwenxiu,yangchengxi\}@sensetime.com }
}


\maketitle
\ificcvfinal\thispagestyle{empty}\fi

%%%%%%%%% ABSTRACT
\begin{abstract}
%
% Time-of-Flight (ToF) sensors measure depth by the reflection time of pulsed near-infrared signal sent to scenes. 
% A ToF sensor is typically positioned side-by-side with an RGB camera capturing color images, toghther they form a ToF RGB-D module. 
% Recently, it is increasingly popular to equip mobile devices with ToF RGB-D modules to support a wide range of 3D applications.
% Since 1) the depth image is collected from the perspective of the ToF sensor and 2) the pre-calibrated camera parameters are versatile during deployment, the obtained depth needs to be rectified on-the-fly to make it well aligned withe RGB image. 
% Moreover, depth measurements of ToF cameras are erroneous and noisy by their nature, which requires further refinement. 
Recently, it is increasingly popular to equip mobile RGB cameras with Time-of-Flight (ToF) sensors for active depth sensing. 
However, for off-the-shelf ToF sensors, one must tackle two problems in order to obtain high-quality depth with respect to the RGB camera, namely 1) online calibration and alignment; and 2) complicated error correction for ToF depth sensing. 
In this work, we propose a framework for jointly alignment and refinement via deep learning. 
First, a cross-modal optical flow between the RGB image and the ToF amplitude image is estimated for alignment. 
The aligned depth is then refined via an improved kernel predicting network that performs kernel normalization and applies the bias prior to the dynamic convolution. 
To enrich our data for end-to-end training, we have also synthesized a dataset using tools from computer graphics. 
Experimental results demonstrate the effectiveness of our approach, achieving state-of-the-art for ToF refinement.
\end{abstract}

%%%%%%%%% BODY TEXT
\setlength{\belowcaptionskip}{-6pt}
\setlength{\textfloatsep}{12pt plus 1.0pt minus 2.0pt}
\setlength{\intextsep}{8.0pt plus 2.0pt minus 2.0pt}
\setlength{\floatsep}{8pt plus 1.0pt minus 2.0pt}
% \titlespacing{\paragraph}{%
%   0pt}{%              left margin
%   1pt}{% space before (vertical)
%   0pt}%  

\section{Introduction}\label{sec:intro}
Nowadays, RGB-D camera modules based on Time-of-Flight (ToF) sensors are becoming increasingly popular for mobile devices. 
At an affordable cost, it provides portable active depth measurements.
In general, compared to monocular or stereo camera modules \cite{godard2017unsupervised,hirschmuller2007stereo,luo2018single,pang2017cascade,pang2018zoom}, ToF sensors provide higher precision depth values for short-range distance sensing \cite{hansard2012time}.
%In general, for short-range distance sensing, ToF sensors provide higher precision depth values than monocular or stereo camera modules \cite{hansard2012time}. 
However, off-the-shelf ToF RGB-D camera modules have two problems: 
\begin{enumerate}[(i)]
    \item {\bf Perspective difference}: The depth measurements are initially defined from the perspective of the ToF sensor, thus alignment between the depth images and RGB images is necessary;
    \item {\bf Erroneous measurements}: depth measurements of ToF sensors suffer from different types of error such as multi-path interference, noise, {\it etc}. 
\end{enumerate}
These two problems hamper the direct usage of ToF RGB-D camera modules for applications such as computational photography, augmented reality and video entertainment.

Multi-view geometry sheds light on the first problem. 
In fact, pixel correspondences between the RGB image and the ToF amplitude image can be computed given the \emph{true depth} from the perspective of either of the images accompanied with the \emph{full set of camera parameters} \cite{hartley2003multiple}. 
% In general, knowledge of any of the two can be used to solve for the third. 
However, under dynamic changes during deployment, mobile ToF RGB-D camera parameters can seldom be calibrated once and for all. 
In fact, modern RGB cameras are often equipped with optical image stabilization (OIS) systems which dynamically changes the principal points, alongside with other mild calibration degradation to the ToF RGB-D camera module. 
These impacts can be sufficiently modeled by the changes of the principal point $c_x$, $c_y$ of the RGB camera, and the relative translation parameters $t_x$, $t_y$ \cite{cardani2006optical,xiao2018dsr}; while the rest of the parameters can be viewed as unchanged.
%In fact, on many modern RGB cameras, the optical image stabilization (OIS) system and the auto-focus system are two major elements changing the initial extrinsic camera parameters. 
% while unintentional collisions may also affect the camera parameters.
% Specifically, the principal point parameters $c_x$, $c_y$ of the RGB camera, and the relative translation parameters $t_x$, $t_y$ can be dynamically changes by the OIS and the auto-focus systems; while the rest of the parameters stay unchanged \red{[cite]}.
Hence, it brings the need of performing online calibration and alignment for ToF RGB-D camera modules.

\begin{figure}
    \centering
    % \subfloat[RGB.]{\includegraphics[width=1.5\widthdef]{figures/pipeline/0103_R.png}} \hspace{\inter}
    % \subfloat[RGB.]{\includegraphics[width=1.2\widthdef]{figures/pipeline/0103_L.png}} \hspace{\inter }
    \subfloat[\scriptsize Unaligned erroneous depth image.]{\includegraphics[width=2\widthdef]{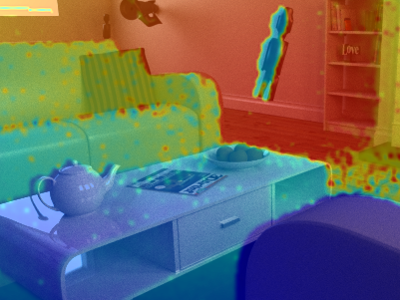}}  \hspace{5\inter}
    \subfloat[Our result.]{\includegraphics[width=2\widthdef]{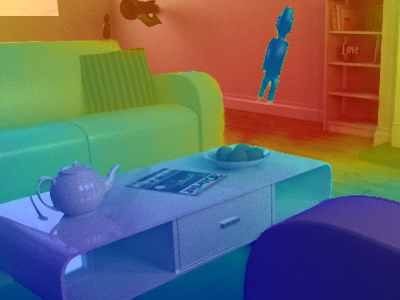}}
    \caption{Proposed framework of alignment and refinement of ToF depth images for weakly calibrated ToF RGB-D module. The scene is chosen from our synthetic ToF-FlyingThings3D dataset.}
    \label{fig:intro}
\end{figure}
%sufficient to model these random changes on only a subset of the parameters---{\it i.e.}, the two principal point parameters $c_x$, $c_y$ of the RGB camera, and the two relative translation parameters $t_x$, $t_y$---while having the same focal lengths $f_x$, $f_y$ after initial calibration when the images are resized to have the same height and width.
With the above practical setup, we assume the ToF sensor and the RGB camera have already been calibrated with standard procedure, {\it e.g.}, with \cite{zhang2000flexible}, and therefore having known initial camera parameters.
However, the set of parameters $\{c_x, c_y, t_x, t_y\}$ changes during deployment.
We call such ToF RGB-D camera modules {\it weakly calibrated}. 
As a result, in the following we also assume both the ToF amplitude images and the ToF depth images provided to our framework {\it have already been rectified and warped} to the viewpoint of RGB camera according to the initial camera parameters,\footnote{From the mechanisms of ToF sensor \cite{hansard2012time}, we note that a ToF amplitude image and its corresponding ToF depth are essentially aligned.} 
However, random perturbations to $\{c_x,c_y,t_x,t_y\}$ lead to misalignment; so performing online alignment is a must.

Although a straightforward solution is to match their key points on the fly, this approach fails in practice because the imaging process of a ToF camera departs greatly from that of a standard RGB camera \cite{hansard2012time}. 
Above all, a ToF amplitude image is lightened up by a single light source located on the module. 
Moreover, since infra-red frequencies are used, the same material may have considerably different appearances in the ToF amplitude images and the color images.
% due to their fundamentally different light sources and spectrum.
% While a standard camera often deploy multiple ambient light sources at visible spectrum for its imaging, a ToF camera images by continuously emitting modulated infra-red light at one or more frequencies, and integrating the reflected light intensity over the exposure time, which typically lasts for a few hundred microseconds. 
% This brings out two main differences between IR and RGB images:
%
%\begin{enumerate}
    %\item[(i)] {%Ambient light must be subtracted from the raw ToF captures, assuming constant ambient light condition during exposure. Hence the acquired IR image will appear to be lightened up by a single light source located at the camera position.
    %The acquired IR image is lightened up by a single infra-red light source located on the module.}
    %\item[(ii)] {Since infra-red frequencies are used for imaging, the same material may have considerably different appearances in its IR and RGB images. %In particular, color textures in RGB may have no correspondence in IR.
    %}
%\end{enumerate}
%
% As results, point matching may be computationally expensive since more global information need to be used, or else it will introduce significant inaccuracy and instability to the alignment problem.

To apply multi-view geometry directly, another difficulty is the second problem---erroneous measurements---as mentioned above.
A ToF sensor approximates the true depth by estimating the phase shift of the received infra-red light, which is determined by the scene geometry, materials, the light source itself, {\it etc}. 
Apart from thermal noise which is common for electronic devices, a major source of error is the multi-path interference (MPI)---stems from the mechanisms of ToF sensor---making the depth measurements farther than the actual ones \cite{hansard2012time}. 
% If we were to adopt the existing point-matching approach, in principle we must refine the depth image before aligning it to the RGB image.
% Although previously there exist ways to reduce errors for depth images from ToF sensors \cite{gupta2015phasor,achar2017epipolar}, their successes are often bound to simple scenes or not yet available for off-the-shelf ToF cameras.

Given the coupled nature of the alignment and the refinement problems, it will be beneficial to solve them with the help from high-quality ToF RGB-D data.
In this paper, we propose a novel end-to-end deep learning framework solving both the alignment and refinement tasks of depth images produced by off-the-shelf ToF RGB-D modules.
Our key contributions include:
\begin{enumerate}
    \item[(i)]{To address the alignment problem, we propose an effective two-stage method for estimating the cross-modal flow between the ToF amplitude and RGB image, utilizing the original depth measurements, and trained with dedicated data augmentation technique.}
    \item[(ii)]{For the ToF depth refinement problem, we propose an effective architecture, \emph{ToF kernel prediction network} (ToF-KPN) which also employs the RGB images. With simple changes to the original KPN, we enable state-of-the-art performance in reducing MPI while enhancing the depth quality.}
    \item[(iii)]{It is difficult to collect sufficient real data with high-quality ground-truth for training. Hence, we \emph{synthesize} a dataset for our problem with tools in computer graphics. We call our dataset ToF-FlyingThings3D, as we let various objects floating in the scenes similar to the FlyingThings3D dataset \cite{MIFDB16}.
    }
\end{enumerate}
%
% We call our \emph{deep end-to-end alignment and refinement} framework DEAR. 
% To the best of our knowledge, we are the first to perform jointly alignment and refinement tasks for ToF RGB-D camera modules via deep learning.

We call our \emph{Deep End-to-end Alignment and Refinement} framework DEAR.
Our paper is organized as follows. 
We review related works in Section\,\ref{sec:related}. 
In Section\,\ref{sec:framework} we elaborate our framework and in Section \ref{sec:data} we detail our data generation and collection strategy. 
Experimentation are presented in Section\,\ref{sec:exp} and conclusions are provided in Section\,\ref{sec:conclude}.
\section{Related Work}\label{sec:related}
To our best knowledge, we are the first in the literature to propose an end-to-end depth alignment and refinement framework for ToF RGB-D camera modules.
Since none of the existing work has the same settings as ours, we briefly review works related to the two components of our framework, namely cross-modal correspondence matching and ToF depth image refinement.
%
%\vspace{-12pt}
\begin{figure*}[htbp]
\centering
    \subfloat[Rough optical flow estimation.]{\includegraphics[height=110pt]{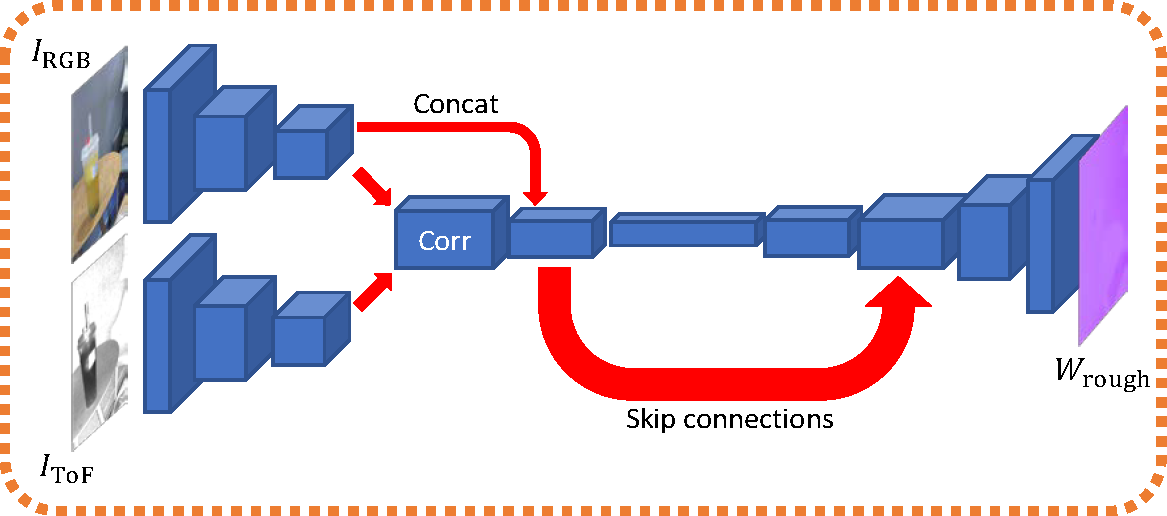}\label{fig:flow_rough}}\hspace{2pt}
    \subfloat[Flow refinement with ToF depth image.]{\includegraphics[height=110pt]{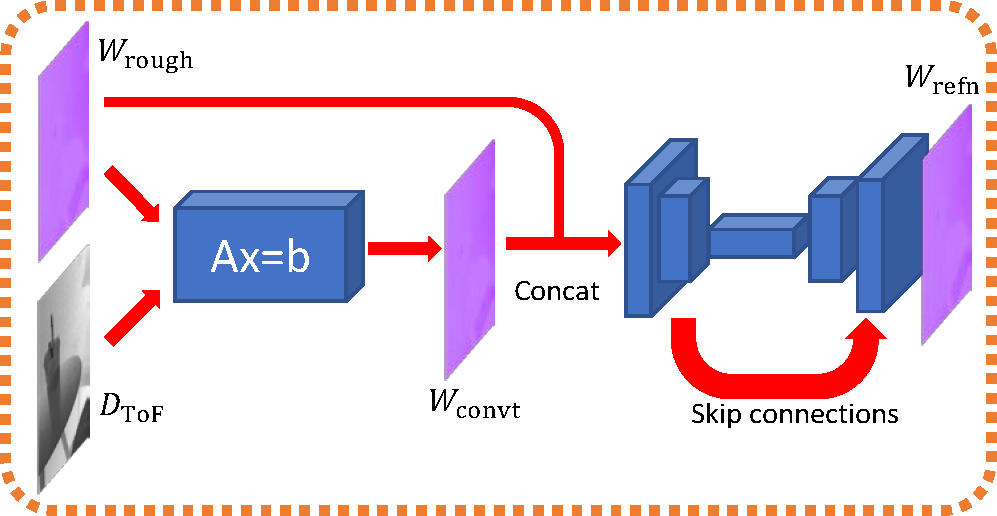}\label{fig:flow_refn}}
\caption{Architecture overview of the cross-modal flow estimation. A rough optical flow is first estimated via FlowNetC. It is then refined by incorporating the depth measurements of the ToF sensor. For flow refinement, we make a depth-flow conversion by estimating the perturbed camera parameters. The converted flow and the rough flow are fed to a small fusion network to obtain the refined flow.}
\label{fig:arch_flow}
\end{figure*}

{\bf Cross-modal correspondence matching.}
Our work performs online cross-modal dense correspondence matching, {\it i.e.}, optical flow estimation, between the ToF amplitude image and the RGB image, so as to address the alignment problem.
In \cite{aguilera2015lghd}, the authors propose the Log-Gabor Histogram Descriptor (LGHD) which adopts multi-scale and multi-oriented Log-Gabor filters to extract feature descriptors from multi-spectrum image pairs, while Shen {\it et al.}~\cite{shen2014multi} exploit the structure variation existing in multi-modal image sets. 
In \cite{chiu2011improving}, Chiu {\it et al.} propose cross-modal stereo for improving the accuracy of Microsoft Kinect \cite{zhang2012microsoft} by combining the three channels of red, green, and blue optimally to mimic the infrared image. 
A very recent work \cite{zhi2018deep} applies a deep neural network for solving the challenging problem of cross-spectral stereo matching using the rectified near infrared and RGB images, where a novel material-aware loss function is proposed specifically for applications in vehicle vision. 
None of the above works takes the ToF amplitude as the alternative modality nor matches correspondence under weakly calibrated stereos. 
Moreover, our method estimates the flow by exploiting the depth image obtained by the ToF sensor while the other works do not take it into account.
%\vspace{-12pt}

{\bf ToF depth image refinement.} 
There exist a notable number of works on mitigating errors of continuous-wave ToF depth images. 
Early works, such as \cite{godbaz2012closed,fuchs2010multipath,freedman2014sra,naik2015light}, often adopt simplified assumptions such as two-path formulation of MPI, leading to closed-form solutions or costly optimization.
Another stream of works focus on the acquisition side, for example using signals in the GHz band instead of the MHz band to mitigate MPI in diffusive environment \cite{gupta2015phasor, kadambi2017rethinking}, or exploiting epipolar geometry of light paths \cite{achar2017epipolar} at the expense of sequential multiple captures.
These methods can produce physically accurate results but are not yet ready for the markets.
Closely related to our methods are the recent works based on deep learning which utilizes physically accurate synthetic data. 
In \cite{marco2017deeptof} an auto-encoder (U-Net) is used to learn the MPI corrected depth directly, while \cite{su2018deep} starts instead from raw correlation measurements aiming for an end-to-end ToF imaging pipeline. Guo~{\it et~al.}~\cite{guo2018tackling} propose deep learning methods that tackle artifacts from multi-frame fusion as well.
All these works are targeted for purely refining the depth images of ToF sensors, so they do not take the corresponding color images into account.
\section{Alignment and Refinement}\label{sec:framework}
%\subsection{Background and problem setting}
This section illustrates our end-to-end framework for joint alignment and refinement. 
Particularly, we first estimate the cross-modal dense optical flow for image alignment, then a novel architecture---ToF kernel prediction network (ToF-KPN)---is proposed for depth refinement. 

\subsection{Cross-Modal Dense Flow Estimation}
We solve the alignment problem by estimating a flow (denoted as $W\in\mathbb{R}^{h\times w\times 2}$) where the RGB image (denoted by $I_{\rm RGB}$) and the ToF amplitude image (denoted by $I_{\rm ToF}$) are regarded as the first and the second images, respectively.
We denote the operation of warping of a one-channel $h\times w$ image $I$ by the flow (a warp field) $W$ as $I_{\rm warped }=I \circ W$, that is,
\begin{equation}\label{eq:warp}
I_{\rm warped }(p)=I\left(m+W_x(p),\,n+W_y(p)\right),
\end{equation}
%
%where we identify the image as a $\mathbb{R}^{c}$-valued function defined on an $h\times w$ grid, and similarly the flow $V$ as a $\mathbb{R}^2$-valued vector field defined on an $h\times w$ grid, and $id$ denotes the identity mapping of this grid.
%This notation also clearly indicates that the kind of warping we use is differentiable with respect to the network parameters, since they can be seen as a form of composition of functions conformed to the same domain using e.g. bilinear interpolation.
where $I_{\rm warped }(p)$ denotes the $p=(m,n)$-th pixel of image $I$, similarly for $I(p)$; and $W_x, W_y\in\mathbb{R}^{h\times w}$ are the $x$- and $y$- components of the estimated optical flow. 
The warping operation as in \eqref{eq:warp} is differentiable with respect to the warp field \cite{jaderberg2015spatial}.
Compared to the classic optical flow estimation approaches, recent approaches via convolutional neural networks (CNNs) not only have strong learning/adaptation power, but are also better at exploiting spatial and non-local information across multiple scales \cite{Ma2019CVPR, dosovitskiy2015flownet}. 
Therefore, we cast the matching task as the estimation of cross-modal dense optical flow with CNNs.
%In other words, we estimate the per pixel correspondence from RGB to IR.
We divide the estimation task into two stages: 1) rough optical flow $W_{\rm rough}\in\mathbb{R}^{h\times w\times 2}$ estimation, and 2) flow refinement.
In the first stage we compute a flow solely based on the $I_{\rm RGB}$ and $I_{\rm ToF}$, while in the second we make use of the depth image of the ToF sensor to refine the flow details.

To compute the rough flow, we have adopted a representative architecture, \emph{FlowNetC} \cite{dosovitskiy2015flownet}, though more advanced choices, {\it e.g.}, PWC-Net \cite{sun2018pwc}, are also applicable. 
FlowNetC is an U-Net with skip connections, where the encoder part contains a Siamese tower followed by a correlation layer computing a cost volume. 
% In details, we first extract common features in RGB and IR using a Siamese tower, followed by a correlation cost volume of these learned features. 
% Afterwards, a U-Net is used to regress the flow from the cost volume.
% The U-Net learns the flows at six scales, with skip connections sharing information across them as well as from the main branch of the feature extractor.
This rough flow estimation module is illustrated in Figure\,\ref{fig:flow_rough}.

%\red{fix}
%From multi-view geometry \red{[cite]}, the pixel correspondence can be computed through
%
%\begin{equation}\label{eq:mvg}
%\begin{pmatrix}x_{RGB}\\
%y_{RGB}
%\end{pmatrix}-\begin{pmatrix}x_{IR}\\
%y_{IR}
%\end{pmatrix}=\begin{pmatrix}\frac{f_{x}t_{x}}{z_{IR}}+c_{x}\\
%\frac{f_{y}t_{y}}{z_{IR}}+c_{y}
%\end{pmatrix},
%\end{equation}
%
%where $(x_{(\Omega)},y_{(\Omega)})$, $\Omega\in\{RGB, IR\}$ are the intrinsic image coordinates in the RGB and IR images for the corresponding scene point, and $z_{IR}$ is ground-truth scene depth on the IR image.
%\red{fix}

In the second stage, we refine the flow by incorporating the depth images obtained by the ToF sensor using a lightweight fusion CNN.
Particularly, we first warp the depth image from the perspective of the ToF camera, denoted by $D_{\rm ToF}$, to the perspective of the RGB camera, $D_{\rm RGB}$, {\it i.e.}, $ D_{\rm RGB} = D_{\rm ToF} \circ W_{\rm rough}$.
For the weakly-calibrated module, we can readily estimate a new set of camera parameters  $\{t_{x}^{\star},t_{y}^{\star},c_{x}^{\star},c_{y}^{\star}\}$ between the ToF amplitude image (after initial rectification) and the RGB image by solving the following least-square problem,\footnote{A detailed derivation of this formulation is presented in the supplementary material.}
\begin{equation}\label{appr}
\{t_{x}^{\star},t_{y}^{\star},c_{x}^{\star},c_{y}^{\star}\}\hspace{-3pt}=\hspace{-3pt}\mathop {\arg\min}\limits_{t_{x},t_{y},c_{x},c_{y}}\hspace{-3pt}\sum_{p}\left\|\hspace{-2pt}W_{\rm rough}(p)\hspace{-3pt}-\hspace{-3pt}\begin{pmatrix}\frac{t_{x}}{D_{\rm RGB}(p)}\hspace{-3pt}+\hspace{-3pt}c_{x}\\
\frac{t_{y}}{D_{\rm RGB}(p)}\hspace{-3pt}+\hspace{-3pt}c_{y}
\end{pmatrix}\hspace{-2pt}\right\|^{2}.
\end{equation}
Solving this problem is equivalent to solving a linear system, which is differentiable. 
Hence, it is embedded as a component in our refinement network.
Then we can \emph{convert} $D_{\rm RGB}$ to another estimated flow, $W_{\rm convt}$ (subscript ${\rm convt}$ denotes it is converted from the depth image), given by
%The formulation comes from the assumed perturbation model in Equation \eqref{eq:1}, and the best flow is simply
%
\begin{equation}
W_{\rm convt} = \begin{pmatrix}\frac{t_{x}^{\star}}{D_{\rm RGB}}+c_{x}^{\star}\\
\frac{t_{y}^{\star}}{D_{\rm RGB}}+c_{y}^{\star}
\end{pmatrix}.
\end{equation}
Finally we concatenate $W_{\rm rough}$ and $W_{\rm convt}$ and feed them into a lightweight fusion U-Net, which outputs the refined flow $W_{\rm refn}$.
The architecture of this fusion CNN is illustrated in Figure\,\ref{fig:flow_refn}.
Having computed the refined flow $W_{\rm refn}$, it is applied on the input depth for later depth refinement, {\it i.e.}, $D_{\rm ToF}\circ W_{\rm refn}$. 
For convenience, we simply use $D$ to denote the final warped depth, $D_{\rm ToF}\circ W_{\rm refn}$, in the rest of the paper.

\begin{figure}[t]
\centering
%\hspace*{-0.5cm} 
    \includegraphics[width=240pt]{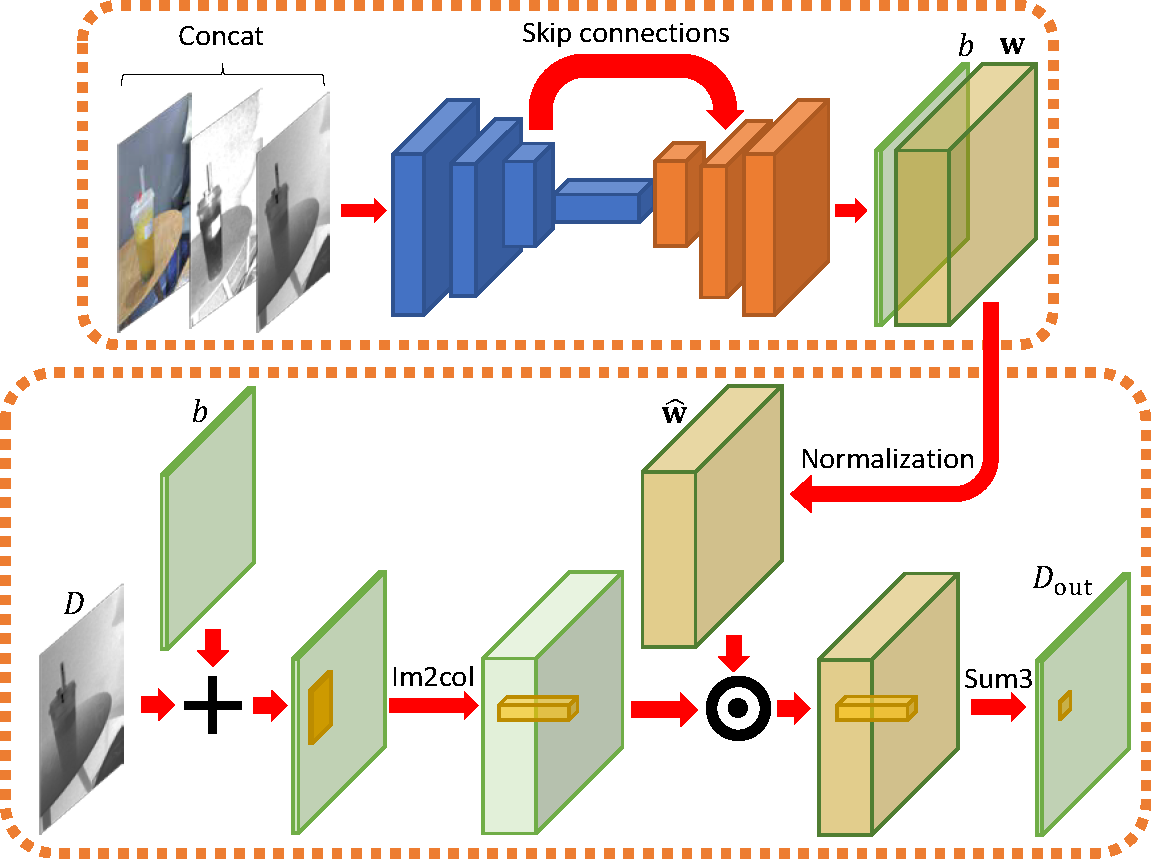}
\caption{Architecture overview of the depth refinement using the proposed ToF kernel prediction network (ToF-KPN). Here ``Im2col'' rearranges each patch along the channel dimension while ``Sum3'' sums along the channel dimension.}
\label{fig:arch_depth}
\end{figure}

\subsection{Refinement via ToF Kernel Prediction Network}\label{ssec:refn}
It is well-known that the ToF depth measurements suffer from error such as the MPI, the ``flying pixel'' artifact, and also thermal noise \cite{hansard2012time}.
Moreover, the warped depth $D$ does not guarantee to be tightly aligned with the RGB image.
Consequently, a post-processing procedure for depth refinement is indispensable.
%It is necessary to refine the depth, $D_{\rm RGB}$, defined from the perspective of the RGB camera, because the original ToF depth measurments suffer from 
% The necessity of such a post-processing step is threefold.
% First of all, so far we have only aligned the ToF depth with the RGB. 
%The MPI error of ToF is yet to be corrected. 
%Secondly, ToF depth is often more blurry than the RGB, which results from the "flying pixels" effect around discontinuous depth, and also from image denoising and resizing inherent in many off-the-shelf ToF cameras.
%Lastly, warping by the refined flow does not always guarantee tight alignment.
% Many depth post-processing pipelines today include an edge-aware filtering stage using a color image as the guidance, such as the fast bilateral solver \cite{barron2016fast, barron2015fast}, \red{more examples}.
% An example is the fast bilateral solver \cite{barron2016fast, barron2015fast}, which can be thought of as finding a smooth approximation of the depth function $D: \mathbb{R}^{h\times w} \to \mathbb{R}$ 
% embedded using RGB image into the 5D {\it bilateral space}. 
% In other words, the domain of the image is transformed from 2D grid points to a set of points in 5D.  
% Edge-aware filtering can be though of as a diffusion process in the bilateral, or similar geometric spaces.
Kernel prediction network (KPN) is a recently proposed model which performs edge-aware adaptive filtering to images in a data-driven manner \cite{Bako2017KPCN,Mildenhall2018Burst,vogels2018denoising}. 
Given depth image $D$, a vanilla (original) KPN uses an U-Net with skip connections to predict for each pixel a kernel operating only on its surrounding patch. 
Specifically, for a KPN with output kernel size $k$ ($k=3$ is used in our work),
\begin{equation}\label{eq:kpn_vanilla}
    D_{\rm out}(p) = {\bf w}^{\rm T}_p \cdot {\rm patch}(D(p)) + b(p),
\end{equation}
where $D_{\rm out}$ is the output depth and $D_{\rm out}(p)$ is its $p$-th pixel, ${\rm patch}(D(p))\in\mathbb{R}^{k^2}$ denotes the vectorized patch of $D$ centered at pixel $p$. The pixel-wise kernel ${\bf w}_p\in\mathbb{R}^{k^2}$ and the bias $b\in\mathbb{R}^{h\times w}$ are outputs of the KPN.
In other words, the KPN output is a 3-D volume of size $h\times w\times (k^2 + 1)$.
We will present an improved KPN for ToF depth image refinement, which differs from \eqref{eq:kpn_vanilla} in two major perspectives. 

First, we empirically find that, in the depth refinement task the vanilla KPN inclines to produce kernel ${\bf w}_p$ with very small magnitudes.
In such cases, \eqref{eq:kpn_vanilla} degenerates to $D_{\rm out} \approx b$ and the KPN behaves like an U-Net. 
To make full use of the filtering of KPN, we normalize the kernel weights by their sum of absolute values, {\it i.e.},
\begin{equation} \label{eq: kernel_norm}
{\widehat {\bf w}_p}(i) = {{{{\bf w}_p}(i)} \mathord{\left/
 {\vphantom {{{{\bf w}_p}(i)} {\sum\nolimits_{i = 1}^{{k^2}} | {{\bf w}_p}|}}} \right.
 \kern-\nulldelimiterspace} {\sum\nolimits_{i = 1}^{{k^2}} | {{\bf w}_p}(i)|}},
\end{equation}
where ${\bf w}_p(i)$ is the $i$-th entry of ${\bf w}_p$. 
% All the $\widehat{w}_i$'s together form the normalized kernel $\widehat{{\bf w}}_i$.

Secondly, resolving MPI is challenging, since it introduces gross error almost uniformly in large area and can hardly be resolved by filtering. 
Consequently, we propose to add the bias term $b(p)$ firstly aiming at correcting the MPI, then use the kernel $\widehat{{\bf w}}_p$ for edge-aware filtering:
\begin{equation}
    D_{\rm out}(p) = {\bf {\widehat w}}_p^{\rm T}\cdot {\rm patch}([D+b](p)),
\end{equation}
%
% The CNN for predicting the biases and kernels is based on a lightweight U-Net with skip connections.
where ${\rm patch}([D+b](p))$ denotes the patch on $D+b$ centered at pixel $p$.
We call our improved KPN as \emph{ToF-KPN} since it is designed for ToF depth image refinement. 
It takes as inputs the RGB image $I_{\rm RGB}$, the warped ToF amplitude image $I_{\rm ToF}\circ W_{\rm refn}$, and the warped depth $D$, and outputs the parameters for elementwise filtering on $D$.
Its filtering scheme is illustrated in Figure\,\ref{fig:arch_depth}.
%However, note that using merely diffusion can never resolve MPI, since MPI introduces gross error almost uniformly in large area and can hardly be averaged out locally by a predicted kernel. 
%Therefore, it is natural to learn an bias term that is aims to correct the MPI along with other errors that are hard to be averaged out locally. 
%The details of the filtering scheme using KPN are as follows. Let $w, b$ denote the predicted spatially-varying kernel and bias by the KPN. We first apply the bias term, and then apply the normalized predicted kernel. Formally, let $\mathcal{P}$ denotes the operation that aggregates the surrounding patch of each pixel $p$ into the channel dimension, and let $s_p$ denote the patch size. 
%The predicted kernels $w$ have the same height and width with the depth image and $s_p^2$ channels. They are first normalized by dividing the absolute sum
%\begin{equation} \label{eq: kernel_norm}
%\hat{w_i} = \frac{w_i}{\sum_{i}^{s_p ^2}|\hat{w}_i|},
%\end{equation}
%then we perform
%\[
%filt\_D = \sum_{i=1}^{s_p^2} \hat{w}_i \cdot (\mathcal{P}(D + b))_i
%\]
%where the subscript index is for the channel dimension. The CNN for predicting the biases and kernels is based on a lightweight U-Net with skip connections. The overall filtering scheme of KPN is illustrated in Figure \ref{fig:arch_depth}.
We have performed extensive ablation studies and will discuss the effects of our modifications in Section \ref{sec:ablation_kpn}. 
These simple changes can boost the results over the vanilla KPN by a \emph{significant} margin.
%Interestingly, we found that enlarging the kernel size or performing normalization on ${\bf w}_i$ lead to a more interpretable bias term \red{not clear}. 
%Otherwise the KPN behaves more similar to a U-Net that regress the depth directly, leading to worse performance.   

\subsection{Loss Functions}
In our work, the training data consists of both the synthetic data with perfect ground-truth and the real data. 
To achieve robustness in both flow estimation and depth refinement, we apply $\ell_{1}$ loss averaged over the image size for training.

{\bf Cross-modal optical flow estimation.} $\ell_{1}$-loss across multiple scales is used in this module. 
Particularly, we denote the network output at scale $s$ by $W^{(s)}_{\rm \Omega}$ and the corresponding ground-truth by $W^{(s)}_{\rm gt}$, where $\Omega\in\{{\rm rough}, {\rm refn}\}$. Then given a training sample, its associated loss is
\begin{equation}
    L_{\Omega} = \sum\nolimits_{s, p} {\frac{\alpha_s}{N_s}}\left\|W^{(s)}_{\rm \Omega}(p)-W^{(s)}_{\rm gt}(p)\right\|_1.
\end{equation}
Here both $W^{(s)}_{\rm \Omega}(p)$ and $W^{(s)}_{\rm gt}(p)$ are $\mathbb{R}^2$ vectors, $N_s$ denotes the number of pixel of that scale. 
We use the same weighting factor $\alpha_s$ as that of FlowNetC \cite{dosovitskiy2015flownet}.
%The choice of loss function is crucial to learning the MPI-corrected geometry and immunity to irrelevant textures in the RGB image. We next turn to our choice of loss functions and study the effects of different losses in Section \ref{sec:exp}.
%Since the training data consist of high quality synthetic data and relatively noisy real ToF data, we use $\ell ^{1}$ losses $\|\cdot\|_1$ averaged over the image size for robustness in both flow estimation and depth refinement tasks.
%For cross modal flow estimation, a multi-scale $\ell^{1}$-loss is used. 
%More precisely, denote the base/refined flow computed at $2^{s}$-downsampling rate by $\downarrow_{s}V_{base/refn}$ and corresponding downsampled
%ground-truth by $\downarrow_{s}gt\_V$. The loss can be written as 
%
%\begin{equation}
%L_{\Omega}(\theta)=\sum_{s} \, %\alpha_{s}\left\|\downarrow_{s}V_{\Omega}(\theta)-\downarrow_{s}gt\_V\right\|_{1},    
%\end{equation}
%
%where $\Omega \in \{base, refn\}$, $\theta$ denotes network parameters, $\alpha_{s}$ is the weighting factor for each loss at scale $s$. 
%We simply use the same weighting factors at corresponding scales in the original FlowNet paper\cite{fischer2015flownet}.

{\bf Depth refinement.} Choosing proper loss functions are crucial for learning correct geometry without MPI and irrelevant textures from the RGB image. 
$\ell_{1}$ losses on the output depth and its gradients are used in this module. 
Particularly, given the output depth $D_{\rm out}$ and the corresponding ground-truth depth $D_{\rm gt}$, its associated loss is
\begin{equation} \label{eq: depth_loss}
\begin{aligned}
L_{\rm depth} & = \frac{1}{N}\sum\nolimits_{p}\left\|D_{\rm out}(p) - D_{\rm gt}(p)\right\|_{1} \\
& \; \; \; + \lambda \|\nabla D_{\rm out}(p) - \nabla D_{\rm gt}(p)\|_1,
\end{aligned}
\end{equation}
where $N$ is the number of pixels, the gradient is computed with the discrete Sobel operator \cite{sobel19683x3}.
%We find that, appropriate weights of $\lambda$ and $\beta$ let the ToF-KPN to learn correct geometry with minimal MPI while preserving details. 
In our experiments, we set $\lambda = 10$ to let the ToF-KPN learn correct geometry with minimal MPI while preserving details. 
We summed up the three loss functions, $L_{\rm rough}$, $L_{\rm refn}$ and $L_{\rm depth}$ for overall end-to-end training.
% Detailed study about the loss functions is presented in Section\,\ref{sec:exp}.

%For depth refinement, we use weighted $\ell^{1}$-losses on depth and its gradients. 
%The inclusion of gradient loss turns out to be essential, and appropriate weighting can enable the KPN to learn the MPI-corrected geometry while preserving details. 
%We will demonstrate these observations with experiments in Section \ref{sec:exp}.
%Formally, the loss is written as
%\begin{equation} \label{eq: depth_loss}
%\begin{aligned}
%L_{D}(\theta) & =  \beta\left\|D(\theta) - gt\_D\right\|_{1} \\
%& \; \; \; + \lambda \|\nabla D(\theta) - \nabla gt\_D\|_1 ,
%\end{aligned}
%\end{equation}
%where $D_{\theta}$ denotes the network output depth, $\theta$ denotes network parameters. In our experiments we find $\lambda = 10$ and $\beta = 1$ yield the best performance, where the two loss terms are roughly equal to each other during training.  

%
\section{Datasets and Augmentation}\label{sec:data}
% This section describes our synthetic data generation and real data collection procedure. 
% We also present our data augmentation scheme in detailed.

\begin{figure}[t]
\vspace{-5pt}
\centering \scriptsize
        \subfloat{\includegraphics[width=1\widthdef]{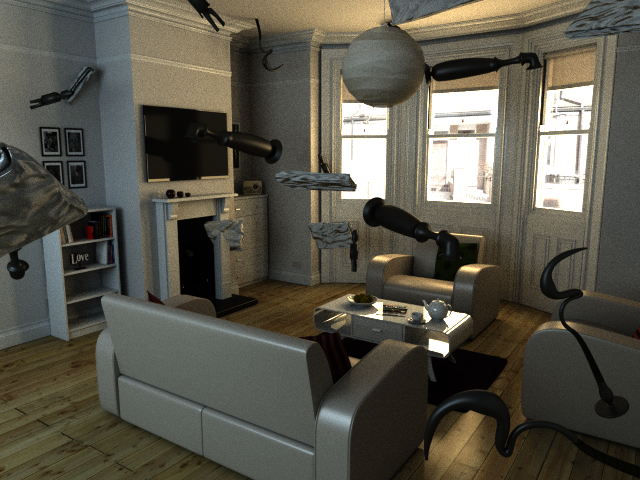}}\hspace{\inter}
        \subfloat{\includegraphics[width=1\widthdef]{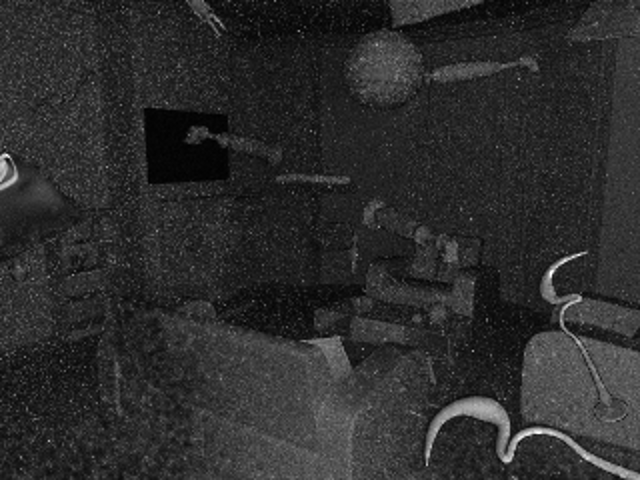}}\hspace{\inter}
        \subfloat{\includegraphics[width=1\widthdef]{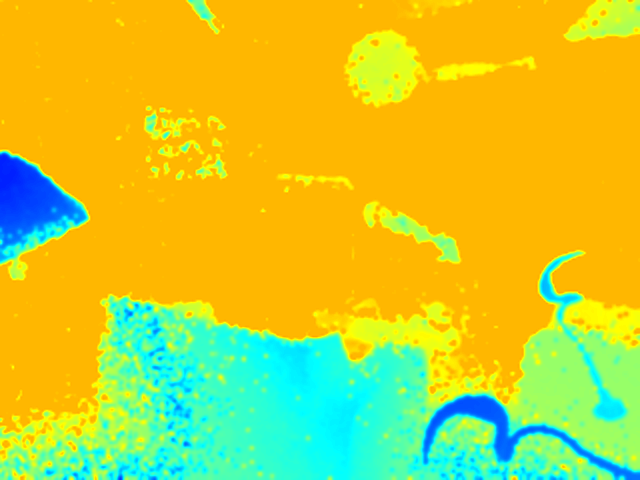}}\hspace{\inter}
        \subfloat{\includegraphics[width=1\widthdef]{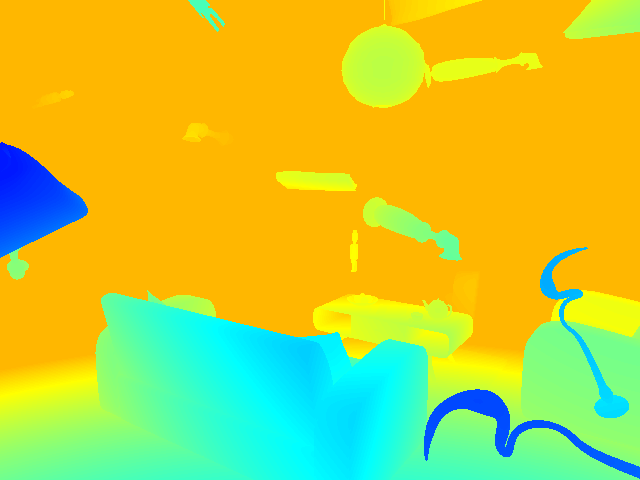}} \\[-2ex]
        \subfloat{\includegraphics[width=1\widthdef]{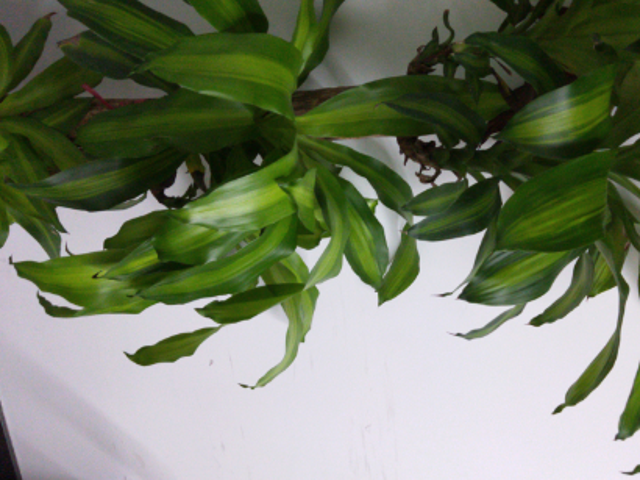}}\hspace{\inter}
        \subfloat{\includegraphics[width=1\widthdef]{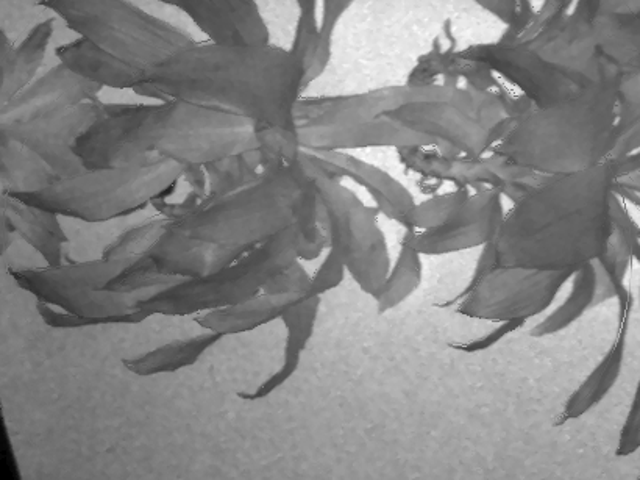}}\hspace{\inter}
        \subfloat{\includegraphics[width=1\widthdef]{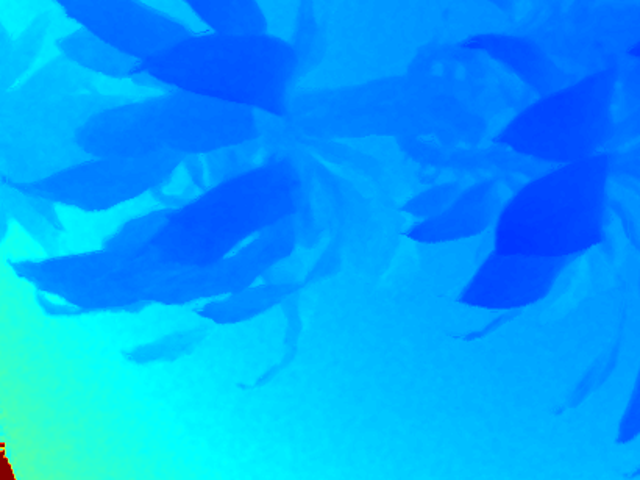}}\hspace{\inter}
        \subfloat{\includegraphics[width=1\widthdef]{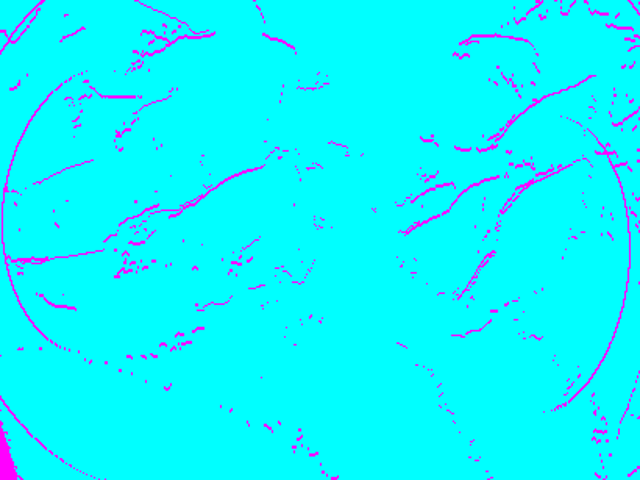}}
 \caption{Examples of our datasets. The first row shows an instance of our synthetic dataset, from left to right are the RGB image, the ToF amplitude, the ToF depth image and the ground-truth depth respectively. The second row shows an instance of our real dataset, from left to right are the RGB image, the ToF amplitude, the ToF depth image and the confidence mask, respectively. We use the cyan color to indicate available pixels on the mask.}
 \label{fig:data}
\end{figure}
\subsection{Synthetic Data Generation}\label{sec:syn}
Due to the mechanisms of ToF depth sensing, it is uneasy to mitigate the error of ToF depth measurements, {\it e.g.}, by using a longer exposure time or a higher modulation frequency \cite{gupta2015phasor,kadambi2017rethinking}.
As a result, collecting a large amount of ground-truth depth images for ToF cameras is very challenging.
Previous works on ToF signal processing  \cite{adam2017bayesian,su2018deep,guo2018tackling,marco2017deeptof} have opt for synthesizing data using {\it transient rendering} from computer graphics \cite{jaderberg2015spatial, smith2008transient}.
We learn from the experience of these previous works to synthesize our dataset.
Technically, we follow the approach provided by Su~{\it et~al}. \cite{su2018deep} in synthetic data generation.
%The dataset of \cite{su2018deep} includes 1250 unique camera views of the publicly available Blender scenes with simple albedo augmentation.
%Since we are particular interested in correlation between image edges and depth details, which is a feature that other synthetic datasets are typically lack of, we enlarge the original dataset of \cite{su2018deep} by randomly placing diverse kinds of objects of various sizes into the scene besides adding another 5000 views. 
Additionally, we randomly place diverse kinds of objects with various sizes into the publicly available {\tt Blender} scenes, totalling 6250 different views for training our framework.
We place our objects in a way similar to the FlyingThings3D dataset \cite{MIFDB16} designed for optical flow estimation. 
Hence, we call our dataset \emph{ToF-FlyingThings3D}.
% For each chosen camera view point of a scene, its transient images are obtained by applying time-bining of light paths to the rendering process in the {\tt pbrt} renderer \cite{pharr2016physically}.
% More sophisticated methods such as \cite{jarabo2014framework} can also be applied.
% Transient images are then converted into ToF intensities and depths.
We also render the corresponding RGB images using {\tt Cycles} in Blender.
These together form the $\{{\tt ToF\;amplitude, RGB, ToF\;depth}\}$ triplets mimicking the outputs of an off-the-shelf ToF RGB-D camera module.
The corresponding ground-truth depths are obtained from Blender's {\tt Z-pass}. 
Each data sample consists of the ToF amplitude, RGB image, ToF depth image, and the ground-truth depth image, all of size $640\times480$ and generated at the \emph{same} view point. 
We randomly set aside $20\%$ of the data instances for testing while the rest are used for training.
An example of our synthesized data is shown in the first row of Figure\,\ref{fig:data}.
More details about the synthetic dataset can be found in the supplementary material.
% During training, it needs to be augmented by warping both the RGB image and the ground-truth depth to another viewpoint corresponds to that of the RGB camera (Section\,\ref{sec:aug}).

% Let us now briefly explain how to get simulated ToF depth from transient images.
% Denote $(I_{t})_{t=1}^{T}$ to be the transient images of a scene under ToF point light source. 
% The ToF correlation image $C$ at pixel $p$ are obtained by taking the inner product with sinusoidal modulated light $L_{\omega}(t)$ at frequency $\omega$:
% \[
% C(p,L_{\omega}) = \sum_{t=1}^{T} I_t(p) \cdot L_{\omega}(t).
% \]
% Note that $(I_t(p))_{t=1}^T$ is simply the TPSF at $p$.
% The phase angle at $p$ used for depth conversion can then be determined by e.g. taking the argument of the complex number $C(p, \cos_{\omega}) + iC(p, \sin_{\omega})$.
% Note that adding noise to the transient images and the correlation images is also a more faithful way to approximate environmental and sensor noises.

\subsection{Real Data Collection}\label{sec:real}
We have also collected a real dataset with several smart-phones equipped with both an RGB camera and a Panasonic ToF depth sensor \cite{Pana2018}.
Each data sample consists of an RGB image, a ToF amplitude image, a depth image, and a binary mask all of size $640\times 480$. 
The binary mask indicates the locations of the depth measurements of high confidence.
Only depth measurements with high confidence are considered as ground-truth during training.
%We collect a dataset of IR-RGB-D triples using mobile phones equipped with ToF RGB-D cameras.
%Each triple is also associated with a binary mask indicating the location of the missing pixel, which is used when calculating loss. 
% For each camera module, we perform carefully cross-calibration for the ToF camera, and carefully manipulate the smartphone so as to avoid any changes of camera configuration during data collection. 
By carefully calibration during the collection of each data sample, we align the depth image, the ToF amplitude image, the binary mask, and the RGB image to the \emph{same} view point by warping.
Our real dataset includes 400 scenes collected under different illumination, in which there are $42\%$ of the samples belonging to indoor and the rest belonging to outdoor.
These data samples complement the aforementioned synthetic dataset. 
Again, $20\%$ of the real data are reserved for testing while the rest are used for training.
An instance of real data is shown in the second row of Figure\,\ref{fig:data}.

\subsection{Data Augmentation via Multi-view Geometry}\label{sec:aug}
We are now equipped with both synthetic data (Section\,\ref{sec:syn}) and real data (Section\,\ref{sec:real}) in which every data sample is well aligned.
During training for the alignment module and end-to-end training, we generate unaligned training samples from the aligned ones \emph{on the fly}.
In this way we enhance the robustness, by making sure that the unaligned ToF and RGB training data cover as much as possible the permissible perturbations of camera parameters.

The perturbation range is determined from the devices used. Specifically, for each sample, we uniformly sample $c_x$, $c_y$ within $\pm 2.5\%$ of the input image size.
For images of size $640 \times 480$, these perturbations can cause the true alignment to deviate from initial calibration by $20$ pixels or more.
Among all the initial $t_x$'s of our ToF RGB-D camera modules, we denote the one with largest absolute value be $t'_x$, similarly for $t'_y$.
Then we uniformly sample $t_x$ and $t_y$ within $\pm30\%$ of $t'_x$ and $t'_y$, respectively.
With multi-view geometry, we use the generated $\{t_x, t_y, c_x, c_y\}$ to compute the forward optical flow from the view of the ToF sensor to a virtual RGB camera.
With this flow, we warp both the ground-truth depth and the RGB image to the view of the virtual RGB camera, leading to the ground-truth depth and the RGB image for training.
We also compute the ground-truth inverse flow regarding the RGB image as the first image and the ToF amplitude image as the second image.
This inverse optical flow is used as the supervising signal for training the alignment module.
Note that we also update the confidence masks that indicate both the occlusion pixels or invalid depth values due to warping. 
These masks are used in the optimization \eqref{appr} and calculation of losses, where the contributions by the invalid pixels are not considered.
\section{Experimentation}\label{sec:exp}
% This section presents comprehensive experiments on our proposed DEAR framework.

\subsection{Training Specifications}
We have adopted a pre-training strategy for both the alignment and refinement modules.
During pre-training, the alignment module is trained in a stage-wise manner, that is, we first trained the FlowNetC only for rough flow estimation, then we included the flow refinement module, both for 20 epochs. 
In parallel, we pre-trained the ToF-KPN for 40 epochs.
We finally stack the alignment and refinement modules together for overall end-to-end fine-tuning for 10 epochs. 
For all the training, we used the ADAM optimizer \cite{Kingma2015AdamAM} with a batch size of $3$, where images are randomly cropped into size $384\times 512 $.
When training from scratch, the learning rates are set to be $4\times 10^{-4}$, while during overall fine-tuning, the learning rates are set to be $1\times 10^{-5}$.
In both cases, we adopt a staircase decay rate of $0.7$ to the learning rates after every two epochs. 
Our implementation is based on the {\it TensorFlow} framework \cite{abadi2016tensorflow}.
All the models are trained on an Nvidia GTX1080 Ti GPU.
The results reported in Section\,\ref{sec:ablation_kpn} and Section\,\ref{sec:compare_tof} are based on the separately trained alignment and refinement modules, and in Section\,\ref{ssec:dear} the jointly fine-tuned DEAR framework.

\subsection{Ablation Studies}\label{sec:ablation_kpn}
{\bf Flow refinement with fusion network.} 
Camera parameter estimation in Figure\,\ref{fig:flow_refn} acts as an intermediate step bringing raw depth information into flow estimation, together with the fusion network it refines the rough optical flow.
We herein quantitatively evaluate the flow estimation results before and after adding the optical flow refinement, as well as directly using depth as fusion network's input, on both the real and synthetic datasets.
Average end-point error (AEPE) is used as the metric for objective evaluation.

\begin{figure}[t]
\centering \scriptsize
       \captionsetup[subfigure]{labelformat=empty}
%        \subfloat[a]{\includegraphics[width=\widthdef]{figures/0023_D_ori.png}}\hspace{\inter}
        \subfloat{\includegraphics[width=0.65\widthdef]{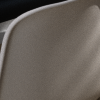}}\hspace{\inter}
        \subfloat{\includegraphics[width=0.65\widthdef]{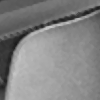}}\hspace{\inter}
        \subfloat{\includegraphics[width=0.65\widthdef]{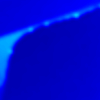}}\hspace{\inter}
        \subfloat{\includegraphics[width=0.65\widthdef]{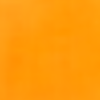}}\hspace{\inter}
        \subfloat{\includegraphics[width=0.65\widthdef]{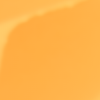}}\hspace{\inter}
        \subfloat{\includegraphics[width=0.65\widthdef]{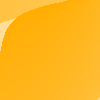}} \\[-3ex]
        \addtocounter{subfigure}{-6}
        \subfloat[RGB]{\includegraphics[width=0.65\widthdef]{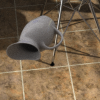}}\hspace{\inter}
        \subfloat[ToF ampl.]{\includegraphics[width=0.65\widthdef]{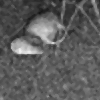}}\hspace{\inter}
        \subfloat[ToF depth]{\includegraphics[width=0.65\widthdef]{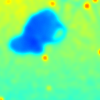}}\hspace{\inter}
        \subfloat[$W_{\text{rough}}$]{\includegraphics[width=0.65\widthdef]{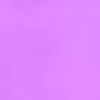}}\hspace{\inter}
        \subfloat[$W_{\text{refn}}$]{\includegraphics[width=0.65\widthdef]{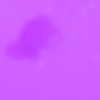}}\hspace{\inter}
        \subfloat[$W_{\text{gt}}$]{\includegraphics[width=0.65\widthdef]{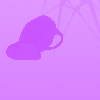}}
\vspace{-8pt}
 \caption{Optical flow refinement incorporating the raw ToF depth measurements greatly refines flow quality.}
 \vspace{8pt}
 \label{fig:pose}
\end{figure}

We first validate the accuracy of our alignment module using both the synthetic data and the real data. 
Specifically, we apply the method described in Section\,\ref{sec:real} to generate test data from randomly sampled camera parameters.
To model different levels of perturbations, we generate 6 groups of data, each containing $1000$ $\{{\tt ToF\;amplitude, RGB, ToF\;depth}\}$ triplets accompanied with the ground-truth flow, where perturbations are sampled from normal distributions with increasing standard deviations.
Our experiments found that the flow refinement module consistently leads to improved accuracy (Table\,\ref{tab:ablation_flow_aepe}).
We also qualitatively demonstrate the effect of flow refinement in Figure \ref{fig:pose}. 

% \begin{table}[t]
%   \centering \footnotesize  
%     \setlength{\tabcolsep}{4pt}
%     \begin{tabular}{l|ab|ab|ab}
%     \hline
%     {Standard Deviation} $\sigma$ & \multicolumn{2}{c}{ 2.00} & \multicolumn{2}{c}{ 4.00} & \multicolumn{2}{c}{ 6.00}  \\
%     \hline
%     {Datasets} & Real & Syn. & Real & Syn. & Real & Syn.\\
%     \hline
%     %   \rowcolor{Gray}
%       {Before Refinement}  & 2.08  & 0.77  & 2.63  & 0.92  & 3.01  & 1.46  \bigstrut[t]\\
%     %   \rowcolor{Gray} {w/ Pose Var}  & {\bf 1.11}  & {\bf 1.07}  & {\bf 1.05}  & {\bf 2.57}  & {\bf 1.54}  & {\bf 4.31} \bigstrut[t]\\
%     {After Refinement} & {\bf 1.99}  & {\bf 0.60}  & {\bf 2.53}  & {\bf 0.75}  & {\bf 2.84}  & {\bf 1.26}\bigstrut[b]\\
%     % {w/o Pose Variance} & 1.15  & 1.12  & 1.26  & 2.73  & 1.67  & 4.72 \bigstrut[b]\\
%     \hline
%     \end{tabular}%
%      \caption{Average end-point-error before and after flow refinement.}
%   \label{tab:ablation_flow_aepe}%
% \end{table}%

\begin{table}[t]
  \centering \footnotesize  
    \setlength{\tabcolsep}{4pt}
    \begin{tabular}{l|ab|ab|ab}
    \hline
    {Standard Deviation} $\sigma$ & \multicolumn{2}{c}{ 2.00} & \multicolumn{2}{c}{ 4.00} & \multicolumn{2}{c}{ 6.00}  \\
    \hline
    {Datasets} & Real & Syn. & Real & Syn. & Real & Syn.\\
    \hline
    %   \rowcolor{Gray}
      {Before Refinement}  & 1.28  & 1.52  & 1.48  & 2.10  & 1.59  & 2.70  \bigstrut[t]\\
      {Direct Fusion}  & 1.29  & 1.55  & 1.50  & 2.16  & 1.63  & 2.79  \bigstrut[t]\\
    %   \rowcolor{Gray} {w/ Pose Var}  & {\bf 1.11}  & {\bf 1.07}  & {\bf 1.05}  & {\bf 2.57}  & {\bf 1.54}  & {\bf 4.31} \bigstrut[t]\\
    {After Refinement} & {\bf 1.15}  & {\bf 1.34}  & {\bf 1.31}  & {\bf 1.87}  & {\bf 1.36}  & {\bf 2.45}\bigstrut[b]\\

    % {w/o Pose Variance} & 1.15  & 1.12  & 1.26  & 2.73  & 1.67  & 4.72 \bigstrut[b]\\
    \hline
    \end{tabular}%
    \vspace{-3pt}
     \caption{Average end-point error before and after flow refinement.}
  \label{tab:ablation_flow_aepe}%
\end{table}%

\begin{figure*}[t]
\vspace{-10pt}
\centering \scriptsize %\captionsetup{justification=centering}
%        \subfloat[a]{\includegraphics[width=\widthdef]{figures/0023_D_ori.png}}\hspace{\inter}
        % \subfloat{\includegraphics[width=1.4\widthdef]{figures/bias_study/0000_R.png}}\hspace{\inter}
        % \subfloat{\includegraphics[width=1.05\widthdef]{figures/bias_study/0000_wD.png}}\hspace{\inter}
        % \subfloat{\includegraphics[width=1.05\widthdef]{figures/bias_study/fD_nonormout_0000.png}}\hspace{\inter}
        % \subfloat{\includegraphics[width=1.05\widthdef]{figures/bias_study/fD_ours_0000.png}}\hspace{\inter}
        % \subfloat{\includegraphics[width=1.05\widthdef]{figures/bias_study/0000_gtD.png}}
        % \subfloat{\includegraphics[width=0.25\widthdef]{figures/cbar_depth.png}}
        % \subfloat{\includegraphics[width=1.05\widthdef]{figures/bias_study/b_nonormout_0000.png}}\hspace{\inter}
        % \subfloat{\includegraphics[width=1.05\widthdef]{figures/bias_study/b_ours_0000.png}}\hspace{\inter}
        % \subfloat{\includegraphics[width=0.25\widthdef]{figures/cbar_bias.png}}
        % \\ \vspace{-10pt}
        % \addtocounter{subfigure}{-9}
        \subfloat[RGB image]{\includegraphics[width=1.15\widthdef]{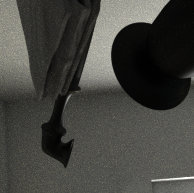}}\hspace{\inter}
        \subfloat[ToF depth]{\includegraphics[width=1.15\widthdef]{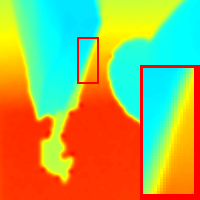}}\hspace{\inter}
        \subfloat[Vanilla KPN]{\includegraphics[width=1.15\widthdef]{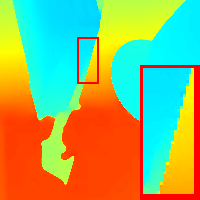}}\hspace{\inter}
        \subfloat[\hspace{-2pt}\textsc{ToF-KPN}(ours)]{\includegraphics[width=1.15\widthdef]{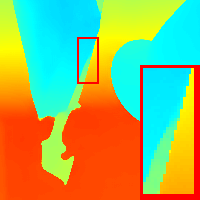}}\hspace{\inter}
        \subfloat[Ground-truth]{\includegraphics[width=1.15\widthdef]{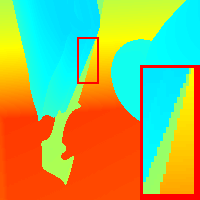}}\hspace{\inter}
        % \subfloat{\includegraphics[width=0.26\widthdef]{figures/cbar_depth.png}}
        % \addtocounter{subfigure}{-1}
        \subfloat[Bias of (c)]{\includegraphics[width=1.15\widthdef]{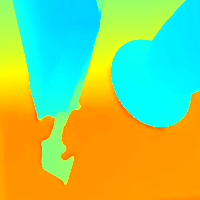}}\hspace{\inter}
        \subfloat[Bias of ours]{\includegraphics[width=1.15\widthdef]{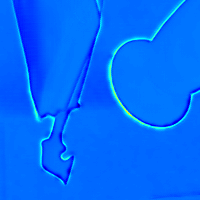}}
        \subfloat{\includegraphics[width=0.28\widthdef]{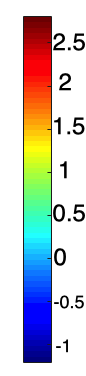}}
       
 \caption{Depth refinement results of an image fragment. The vanilla KPN, {\it i.e.}, \textsc{NoNormAftBias} in (c), produces dominating bias term and diminishing kernels, which behaves very close to a simple U-Net. As shown in (f), the bias image is very similar to the depth itself. In contrast, our approach produces well-behaved bias image (g).}
 \label{fig:bias}
\end{figure*}

{\bf Depth refinement with ToF-KPN.} 
Recall that for depth refinement, we aim to not only enhance the depth details by exploiting the RGB image, but also reduce the ToF depth sensing error such as the MPI and the sensor noises.
This experiment shows that superior refinement quality can be achieved with our proposed ToF-KPN architecture. %by properly training our ToF-KPN architecture on the generated ToF-FlyingThings3D dataset.
Specifically, we validate the performance of our refinement module, denoted by \textsc{ToF-KPN}, against several networks and hyper-parameter variations, they are:
\begin{itemize}\small
    \item \textsc{U-Net}: A U-Net with the same structure as the backbone of our \textsc{ToF-KPN}, but instead it directly regresses the depth. It is supervised using the same loss function \eqref{eq: depth_loss} as the \textsc{ToF-KPN}. 
    
    \vspace{-7pt}
    \item \textsc{NoGrad}: The same with \textsc{ToF-KPN} except is trained using no additional gradient loss as compared to \eqref{eq: depth_loss} of \textsc{ToF-KPN}. 
    
    \vspace{-7pt}
    \item \textsc{NoNorm}: The same with \textsc{ToF-KPN} except the kernel normalization step \eqref{eq: kernel_norm} is not performed.
    \vspace{-7pt}
    \item \textsc{AftBias}: The same with \textsc{ToF-KPN} except the bias is added after applying the kernel.
    \vspace{-7pt}
    \item \textsc{NoNormAftBias}: The same with \textsc{NoNorm} except the bias is added after applying the kernel, {\it i.e.}, the vanilla KPN as in \eqref{eq:kpn_vanilla}.
    \vspace{-7pt}
    % \item \textsc{NoNormKrn5}: The same with \textsc{NoNormAftBias} except the kernel size of the KPN is $5$ instead of $3$.
    % \vspace{-5pt}
    % \item \textsc{NoNormMul}: The same with \textsc{NoNormAftBias} except the kernels and biases are regressed at three scales and on the first two smaller scales they are supervised using  $\ell^{1}$ loss on the filtered depth.
    % \vspace{-5pt}
    \item \textsc{NoNormNoBias}: The same with \textsc{NoNorm} except that no bias term is added.
\end{itemize}
We follow the experimentation approach as in \cite{adam2017bayesian,su2018deep} to analyze the model behaviors. 
Specifically, we sort the pixel-wise errors between the input depth and the ground-truth depth within range of $4$ meters in ascending order and divide them into four quantiles, by which the pixels are classified.
The first quantile ($0 \sim 25\%$) consists of the pixels that are identified as having low-error, while the second ($25 \sim 50\%$) and the third ($50 \sim 75\%$) quantiles are mid- and high-error pixels. 
Errors in the last quantile are treated as outliers.
On the test split of our synthetic ToF-FlyingThings3D dataset, we compute the overall MAE as well as the MAEs of individual classes, and report them in Table\,\ref{tab:design_depth}.

\begin{table}[t]
  \centering\footnotesize
    \begin{tabular}{c|ccc|c}
    \hline
    \multirow{2}{*}{Model} & \multicolumn{4}{c}{ Mean Absolute Error (MAE) in cm}  \\\cline{2-5}
     &  Low Err.     & Mid Err.     & High Err.     & All \bigstrut\\
    \hline
    \rowcolor{Gray} {\textsc{U-Net}}  & { 1.71}  & { 1.42}  & { 1.52}   & { 1.79} \bigstrut[t]\\
    {\textsc{NoGrad}}  & { 2.19}  & {1.78}  & { 1.96}  & { 2.43}  \bigstrut[t]\\
    \rowcolor{Gray} 
    {\textsc{NoNorm}}  & 1.60 &	1.37 &	1.51 &	1.73   \bigstrut[t]\\
    {\textsc{AftBias}} & 1.52 &	1.29 &	1.39 &	1.62 \bigstrut[t]\\
    \rowcolor{Gray}
    % {\textsc{NoNorm5}} & 1.58 &	1.35 &	1.49 &	1.72   \bigstrut[t]\\
    % {\textsc{NoNormMul}} & 1.62 &	1.34 &	1.46 &	1.72    \bigstrut[t]\\
    \rowcolor{Gray}
    {\textsc{NoNormAftBias}} & 1.64 &	1.38 &	1.52 &	1.76  \bigstrut[t]\\
    {\textsc{NoNormNoBias}} & 1.63 &	1.37 &	1.50 &	1.74   \bigstrut[t]\\
    \rowcolor{Gray}
    {\textsc{ToF-KPN} (ours)}  & {\bf 1.44} &	{\bf 1.19} &	{\bf 1.29} &  {\bf 1.51}  \bigstrut[t]\\
    \hline
    \end{tabular}%
    \vspace{-3pt}
    \caption{Quantitative study of model design for the depth refinement module on the ToF-FlyingThings3D dataset.}
  \label{tab:design_depth}%
  \vspace{8pt}
\end{table}%

\begin{table}[t]
  \centering\scriptsize
   \setlength{\tabcolsep}{4pt}
    \begin{tabular}{c|ccc|c|c}
    \hline
    \multirow{2}{*}{\footnotesize Model} & \multicolumn{4}{c|}{Mean Absolute Error (MAE) in cm} & \multirow{2}{*}{No. of Param.} \bigstrut\\\cline{2-5}
    &  Low Err.    & Mid Err.     & High Err.     & All  &   \bigstrut\\
    \hline
    \rowcolor{Gray} \textsc{DeepToF}\footnotemark \cite{marco2017deeptof}  
    & {4.31}  & {3.52}  & {4.08}   & {4.69} & {{\bf 2.6\,M}}\bigstrut[t]\\
    {Su~{\it{et al.}}} \cite{su2018deep}  
    & {4.58} & {4.14}  & {4.57}  & {4.90} & {24.3\,M} \bigstrut[t]\\
    \rowcolor{Gray} {\textsc{ToF-KPN} w/o RGB}  
    & {\bf 2.21} & {\bf 1.93}  & {\bf 2.21}  &	{\bf 2.44}  & {\bf 2.6\,M}\bigstrut[t]\\
    \hline
    \end{tabular}%
    \vspace{-3pt}
    \caption{Quantitative comparison with competitive ToF depth image refinement methods on the ToF-FlyingThings3D dataset. Note that in this comparison no color images are used as inputs.}
  \label{tab:compare_depth}%
\end{table}%
 \footnotetext{Adopted from \cite{marco2017deeptof}, please refer to the text for details.}
 
%  The greatest improvement is provided by the gradient loss term with appropriate weight, as all other versions compared against \textsc{U-Net} show at least 33.0\% increase in performance. Using kernel normalization and adding bias before applying the kernel together improve over other variants by significant margins ranging from 5.3\% to 15.1\% in all metrics.

We first observe that our \textsc{ToF-KPN} provides the best MAE across all error levels.
By comparing \textsc{ToF-KPN} and \textsc{NoGrad}, we note that the greatest gain comes from the weighted gradient loss, without which it results in at least $60.9\%$ increase in MAE. 
With the same loss functions, different model architectures also result in different performances.
The worst behaving KPN variant is \textsc{NoNormAftBias}, {\it i.e.}, the vanilla KPN \eqref{eq:kpn_vanilla}, which neither have kernel normalization nor add the bias first.
For this model, we empirically find that the bias quickly dominates while the kernels degenerates to zeros during training.
Hence, the network behave very similar to \textsc{U-Net}, as mentioned in Section\,\ref{ssec:refn}. 
To mitigate this phenomenon and fully utilize the power of KPN, one may either use kernel normalization or applying the bias beforehand, leading to slightly smaller MSE (\textsc{AftBias} and \textsc{NoNorm}).
%We first observe that all models behave consistently across different error levels.
%The greatest gain is from appropriate weighted gradient loss term, without which it results in at least $60.9\%$ increase in MAE. 
%Having the same loss, the subtle differences in KPN architectures requires a more careful study.
% The worst behaving KPN variant, \textsc{NoNormBiasAft}, has no explicit kernel normalization and the kernels do not operate on the biases. 
% For this model we found that the bias term quickly becomes dominating during training and the kernel term essentially degenerates, hence the network behaves more like \textsc{U-Net}.
%This phenomenon can be mitigated by adding the bias term before applying the kernel \textsc{NoNorm}, or kernel normalization \textsc{BiasAft}. 
However, we furthermore note that for \textsc{NoNorm}, the bias term has little contribution since its performance is similar to the one without bias term, {\it i.e.}, \textsc{NoNormNoBias}.
Performing both kernel normalization and adding bias in the first place as our \textsc{ToF-KPN} leads to the best performance with a substantial margin of $6.8\%$ over the second best model, \textsc{AftBias}. 
A subjective comparison between \textsc{NoNormAftBias} and \textsc{ToF-KPN} is also shown in Figure \ref{fig:bias}, where \textsc{NoNormAftBias} has dominating bias while our \textsc{ToF-KPN} gives more faithful results.

\begin{figure}[t]
\vspace{-20pt}
\centering \scriptsize
    \captionsetup[subfigure]{labelformat=empty}
    \subfloat[ToF amplitude]{\includegraphics[width=1.7\widthdef]{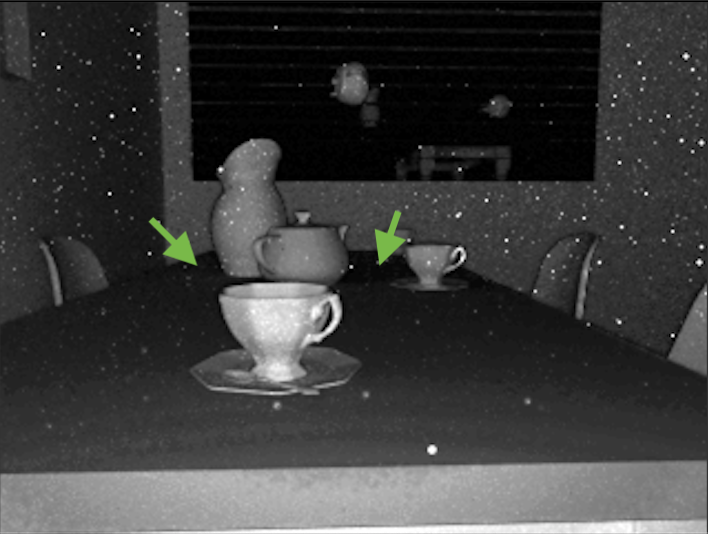}}\hspace{23\inter}
    \subfloat[ToF depth image]{\includegraphics[width=1.7\widthdef]{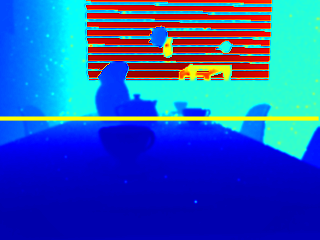}} \\[-2ex]
    \subfloat[ToF depth values] {\includegraphics[width=1.14\widthdef]{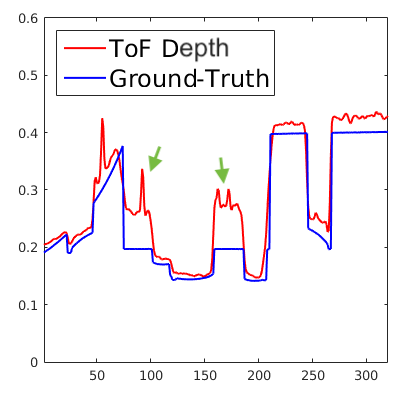}}\hspace{-1pt}
    \subfloat[\textsc{DeepToF}~\cite{marco2017deeptof}] {\includegraphics[width=1.02\widthdef]{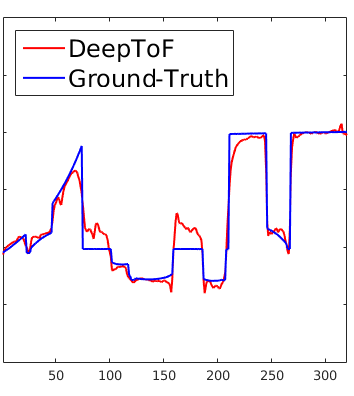}}\hspace{-1pt}
    \subfloat[Su~{\it{et al.}}~\cite{su2018deep}] {\includegraphics[width=1.02\widthdef]{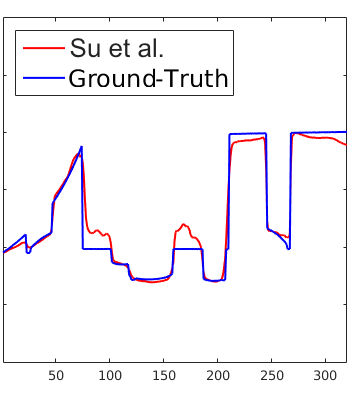}}\hspace{-1pt}
    \subfloat[\textsc{ToF-KPN} (ours)]{\includegraphics[width=1.02\widthdef]{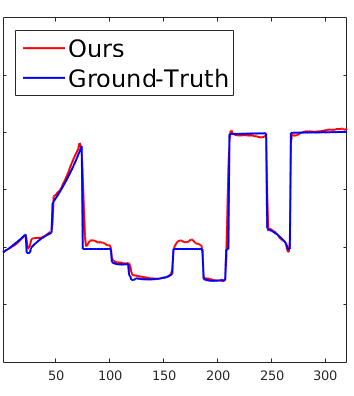}}
 \caption{Depth values of different 
 .0approaches on a scan-line are shown, alongside with the ground-truth. The green arrows indicate the locations that suffer from severe MPI effect.}
 \label{fig:exp_tof}
\end{figure}

\begin{figure*}[t]
%\vspace{-10pt}
\centering \scriptsize
        \subfloat{\includegraphics[width=1.4\widthdef]{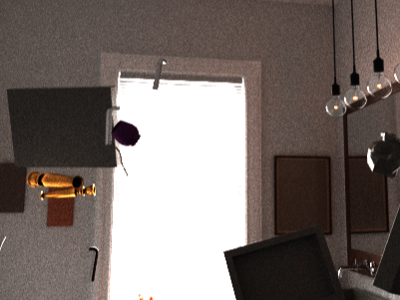}}\hspace{\inter} 
        \subfloat{\includegraphics[width=1.4\widthdef]{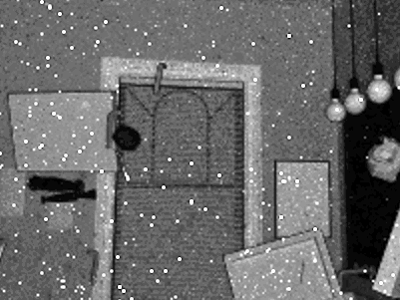}}\hspace{\inter}
        \subfloat{\includegraphics[width=1.4\widthdef]{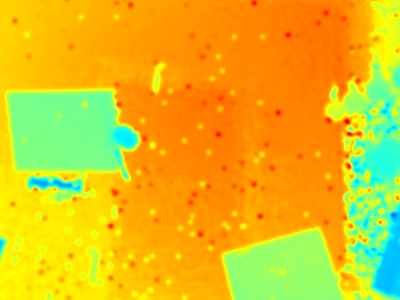}}\hspace{\inter}
        \subfloat{\includegraphics[width=1.4\widthdef]{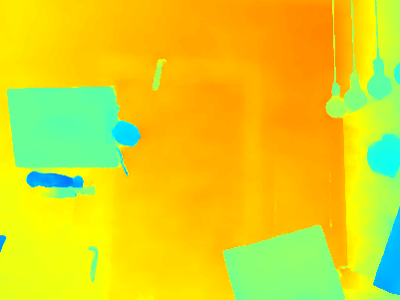}}\hspace{\inter}
        \subfloat{\includegraphics[width=1.4\widthdef]{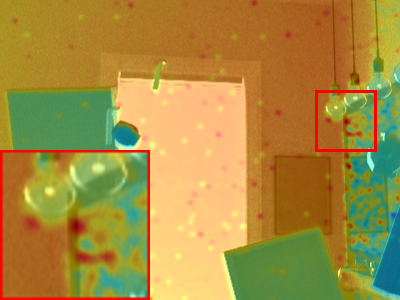}}\hspace{\inter}
        \subfloat{\includegraphics[width=1.4\widthdef]{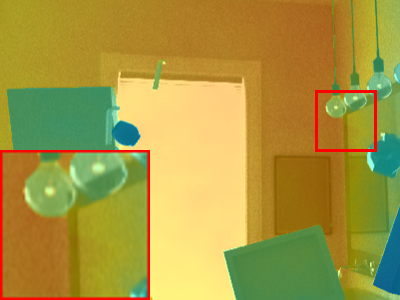}}\\[-3ex]
        \subfloat{\includegraphics[width=1.4\widthdef]{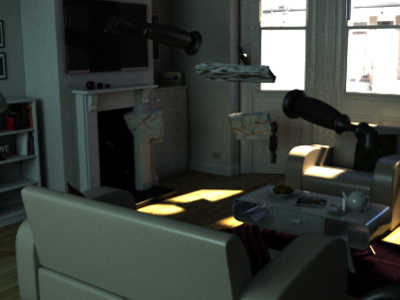}}\hspace{\inter}
        \subfloat{\includegraphics[width=1.4\widthdef]{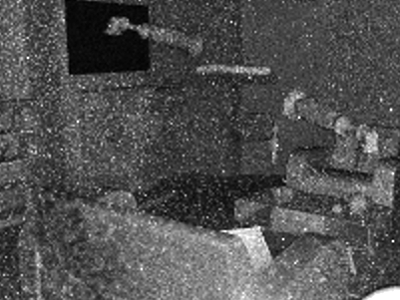}}\hspace{\inter}
        \subfloat{\includegraphics[width=1.4\widthdef]{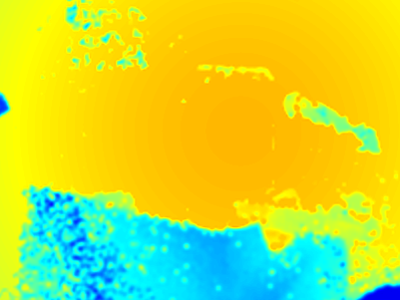}}\hspace{\inter}
        \subfloat{\includegraphics[width=1.4\widthdef]{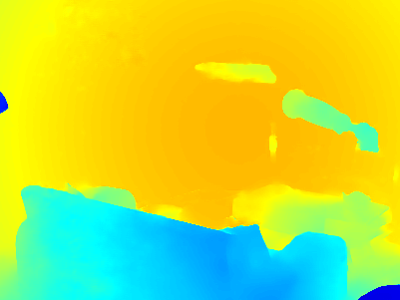}}\hspace{\inter}
        \subfloat{\includegraphics[width=1.4\widthdef]{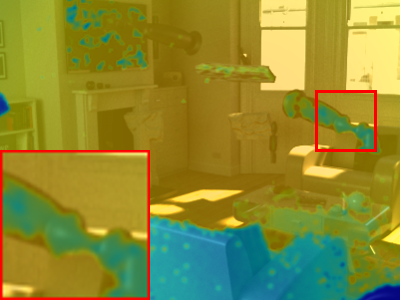}}\hspace{\inter}
        \subfloat{\includegraphics[width=1.4\widthdef]{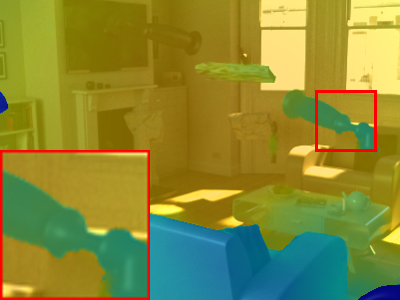}}\\[-3ex]
        \subfloat{\includegraphics[width=1.4\widthdef]{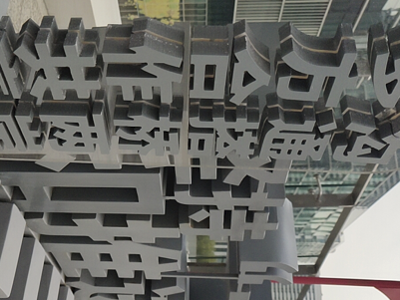}}\hspace{\inter}
        \subfloat{\includegraphics[width=1.4\widthdef]{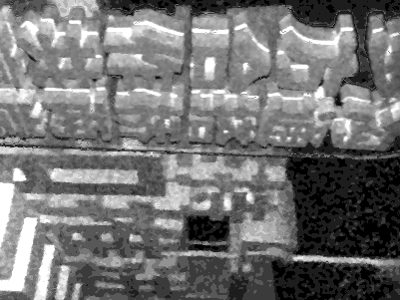}}\hspace{\inter}
        \subfloat{\includegraphics[width=1.4\widthdef]{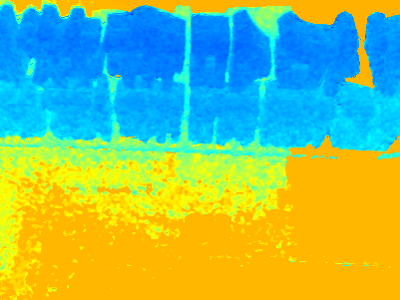}}\hspace{\inter}
        \subfloat{\includegraphics[width=1.4\widthdef]{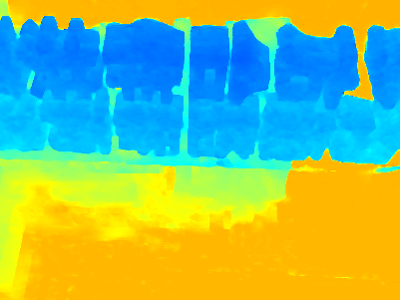}}\hspace{\inter}
        \subfloat{\includegraphics[width=1.4\widthdef]{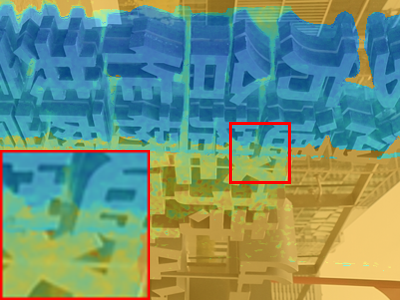}}\hspace{\inter}
        \subfloat{\includegraphics[width=1.4\widthdef]{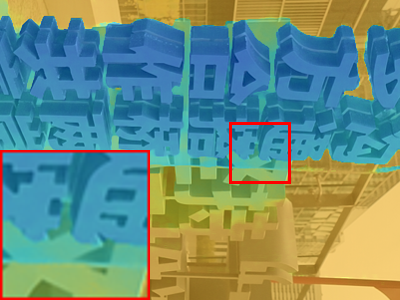}}\\[-3ex]
        \addtocounter{subfigure}{-18}
        \subfloat[RGB image]{\includegraphics[width=1.4\widthdef]{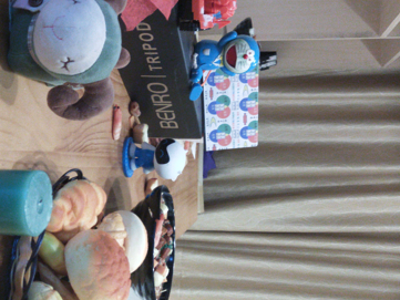}}\hspace{\inter}
        \subfloat[ToF amplitude]{\includegraphics[width=1.4\widthdef]{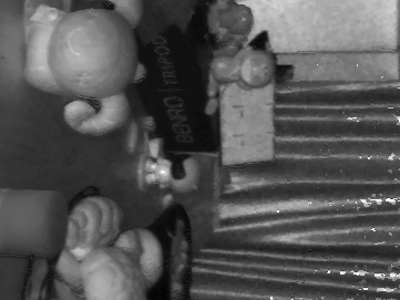}}\hspace{\inter}
        \subfloat[ ToF depth]{\includegraphics[width=1.4\widthdef]{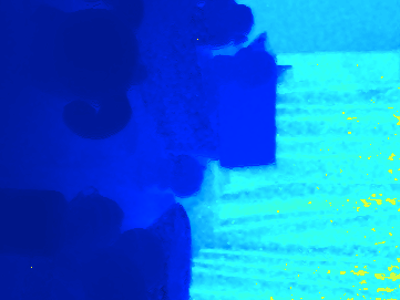}}\hspace{\inter}
        \subfloat[Results of DEAR]{\includegraphics[width=1.4\widthdef]{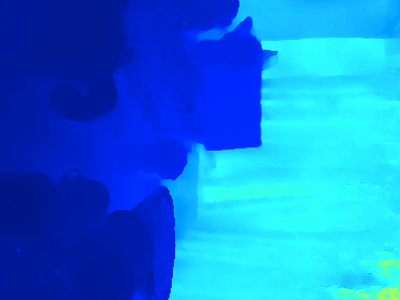}}\hspace{\inter}
        \subfloat[ToF depth + RGB]{\includegraphics[width=1.4\widthdef]{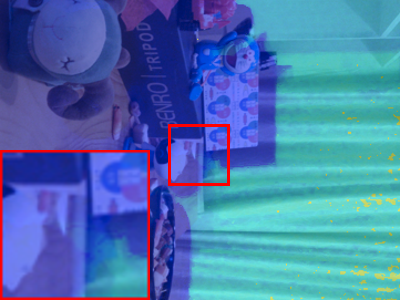}}\hspace{\inter}
        \subfloat[DEAR + RGB]{\includegraphics[width=1.4\widthdef]{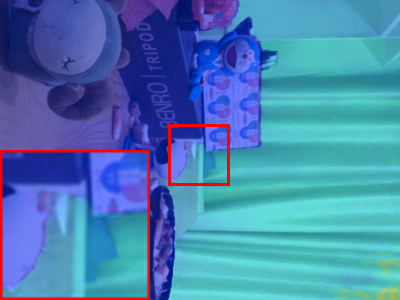}}\\[-2ex]
        
        % \subfloat[RGB.]{\includegraphics[width=1.4\widthdef]{figures/pipeline/0003_R.png}}\hspace{\inter}
        % \subfloat[ToF amplitude.]{\includegraphics[width=1.4\widthdef]{figures/pipeline/0003_L.png}}\hspace{\inter}
        % \subfloat[ ToF depth.]{\includegraphics[width=1.4\widthdef]{figures/pipeline/0003_D.png}}\hspace{\inter}
        % \subfloat[Result.]{\includegraphics[width=1.4\widthdef]{figures/pipeline/0003_fD.png}}\hspace{\inter}
        % \subfloat[ToF depth align vis. .]{\includegraphics[width=1.4\widthdef]{figures/pipeline/com_0003_visD.png}}\hspace{\inter}
        % \subfloat[Result align vis.]{\includegraphics[width=1.4\widthdef]{figures/pipeline/com_0003_visfD.png}} 
    \caption{Visual results of our deep end-to-end alignment and refinement framework. In the first two rows we show the results on synthetic data, while last two rows for real data taken by weakly calibrated ToF RGB-D camera modules.}
 \label{fig:pipeline}
\end{figure*}

\subsection{Comparisons on ToF Depth Image Refinement}\label{sec:compare_tof}
%
% Our synthetic dataset ToF FlyingThings with high quality ground truth makes it possible to compare different ToF depth processing methods quantitatively and more rigorously. 
We compare our proposed ToF-KPN with the state-of-the-art ToF depth image refinement approaches based on deep neural networks.

{\bf Experiments on ToF-FlyingThings3D.}
We compare our proposal with two other representative approaches.
The first one is a deep end-to-end ToF pipeline proposed by Su~{\it{et al.}}~\cite{su2018deep} which takes the raw correlation measurements as inputs.
In the experiment, we directly use their released model because our ToF-FlyingThings3D dataset is generated using the same scenes and settings as \cite{su2018deep}.
The second competing method is the \textsc{DeepToF} framework based on an auto-encoder which processes off-the-shelf ToF depth images directly \cite{marco2017deeptof}.
The original \textsc{DeepToF} employs a model smaller than ours and it is trained on their real dataset.
For fair comparison, we replace their model by our \textsc{U-Net} backbone and train it on our synthetic dataset.
We also apply the Euclidean norm as the loss function as indicated in \cite{marco2017deeptof}.
%: 1) the deep end-to-end ToF pipeline proposed by Su~{\it{et al.}}~\cite{su2018deep} which takes the raw correlation measurements as inputs; and 2) the \textsc{DeepToF} framework based on an auto-encoder which processes off-the-shelf ToF depth images directly \cite{marco2017deeptof}.
Note that these two methods takes as inputs the ToF depth image and the ToF amplitude, {\it i.e.}, they do not use the RGB image.
For fairness, we train a version of our \textsc{ToF-KPN} which does not take the RGB image as input.

%Several remarks on this comparison study are in order.
%First, note that the these methods takes no color image input, therefore for fair comparison we have trained a version of our depth refinement module by simply removing the color image input.
%Second, note that our synthetic dataset is more compatible with that of \textsc{Gan}, so we directly use the model released by Su~{\it{et~al.}} \cite{su2018deep}.
%We modified \textsc{DeepToF} in order for a fair comparison, which in its original form is very slim (max $32$ channels) with more downsampling operations ($2^5\times$spatial decimation), and is supposed to be also trained on real data.
%Hence we have trained on our synthetic training dataset using our \textsc{Baseline} backbone without color image input using an Euclidean norm loss as indicated in \cite{marco2017deeptof}. 
The objective results, in terms of MAE, are presented in Table\,\ref{tab:compare_depth}.
We see that our approach, \textsc{ToF-KPN}, achieves the best performance with minimal amount of model parameters.
In Figure\,\ref{fig:exp_tof}, we demonstrate our capability of reducing  MPI by plotting the depth values along a scan-line. 
% Compared to the results of \textsc{DeepToF} \cite{marco2017deeptof} and Su~{\it{et al.}}~\cite{su2018deep}, our \textsc{ToF-KPN} have not only largely suppressed the MPI (note the locations indicated by the green arrows) and but also provided noise-free depth values.
% Qualitatively, we found that the \textsc{DeepToF} variant tends to be noisier due to only having the Euclidean loss, while \textsc{Gan} is smoother, utilizes multi-frequency correlation to reduce MPI, but fails to retain finer depth details.

{\bf Experiments on FLAT \cite{guo2018tackling}.} We compare our refinement with the multi-reflection module (MRM) in FLAT on 120 static test images provided in the FLAT dataset. 
The MRM uses a KPN architecture but performs filtering on the raw correlation measurements. 
We fine-tune our model on the static training dataset in FLAT, using the depths obtained from the default de-aliasing algorithm used in {\tt libfreenect2} \cite{openkinect_2019} as input. 
Note that we do not train nor test on the images of objects without complete background environment, which have little MPI error but takes up about half of the entire FLAT dataset. 
In testing, we achieve an MAE of $0.68$\,cm while that of MRM is $3.88$\,cm.

\subsection{Evaluation of Deep End-to-End Alignment and Refinement Framework}\label{ssec:dear}
In this last experiment, we evaluate the overall performance of our deep end-to-end alignment and refinement (DEAR) framework on both the synthetic and real datasets. 
For this purpose we generate 150 extra misaligned $\{{\tt ToF\;amplitude, RGB, ToF\;depth}\}$ triplets (accompanied with the ground-truth depth) for testing. 
They are rendered at novel views defined by randomly sampled camera parameters.
The visual results are demonstrated in Figure\,\ref{fig:pipeline}, where the first two rows show results of the synthetic data while the rest show results of our real data.
To visualize the alignment quality, in the last two columns of Figure\,\ref{fig:pipeline}, we blend the RGB images with the corresponding input depth $D_{\rm ToF}$ and the output depth ${D_{\rm out}}$, respectively.

% We have also quantitatively evaluated our performance on the synthetic data given misaligned raw input $\{{\tt ToF\;amplitude, RGB, ToF\;depth}\}$ triplets to our end-to-end system. 
Quantitatively, by assembling the separately trained alignment and refinement modules then applying them to the synthetic data, the average depth MAE reduces from $14.61$\,cm to $2.90$\,cm.
By jointly fine-tuning the overall DEAR framework, the average MAE further reduces to $2.81$\,cm.
%separately trained alignment and refinement modules together \emph{reduces the depth MAE from $8.79$\,cm to $2.90$\,cm with refined flow AEPE $0.72$}, joint fine-tuning  \emph{further reduces depth MAE to $2.81$ cm with refined flow AEPE $0.69$}. 
% on this testing dataset, which is strictly more challenging than the one under the naturally aligned scenario.
This demonstrates that our proposal is capable of producing high-quality refined depths that are also well aligned with the corresponding RGB images.
More results can be found in the appendix.

\section{Conclusion}\label{sec:conclude}
We have proposed DEAR, a deep end-to-end alignment and refinement framework for weakly calibrated ToF RGB-D camera module. 
Our alignment module estimates cross modal optical flow, integrating information from the ToF depth; our refinement module, based on a specifically designed kernel prediction network, tackles the erroneous ToF depth measurements. 
To obtain high-quality data for training we have synthesized a dataset, ToF-FlyingThings3D, with tools from computer graphics. 
Comprehensive experiments have been conducted to demonstrate the effectiveness of our proposal.

\appendix
\appendixpage
\section{More Details on Framework}\label{sec:framework}
In this section, we first derive the formulation of subproblem (2) in the paper via multi-view geometry. 
We then provide the detailed network architectures being used in our work.

\subsection{Derivation of Subproblem (2)}\label{ssec:derivation}
%
%The key is the equation that relates the the second image to the first image, where the second image is taken at the view defined by the camera parameters $t_x, t_y, c_x, c_y$. 
The key of deriving subproblem (2) in the paper is to obtain the relationship of the pixel locations between the first image (the RGB image) and the second image (the ToF amplitude image), where the second image is taken at the viewpoint defined by the camera parameters $\{t_x, t_y, c_x, c_y\}$.
We adopt the simple linear camera model \cite{hartley2003multiple} since we have assumed the weakly calibrated setting.
%Since we assume the weakly calibrated setting, we adopt simple linear camera models \cite{hartley2003multiple}. 
In this regard, we let the world coordinate to be aligned with the first camera, so that the first camera matrix is of the form
\begin{equation}
\mathbf{P}=\mathbf{K}(\mathbf{I}\,|\,\mathbf{0})=\begin{pmatrix}f_{x} & 0 & 0\\
0 & f_{y} & 0\\
0 & 0 & 1
\end{pmatrix}\begin{pmatrix}1 & 0 & 0 & 0\\
0 & 1 & 0 & 0\\
0 & 0 & 1 & 0
\end{pmatrix}.
\end{equation}
We choose the measuring unit to be in pixels. Thus if $\mathbf{x}=(x,y,z)^{\rm T}$ is a scene point in the world coordinate (hence $z$ is the depth with respect to the first camera), its imaged position $(x_{1},y_{1})^{\rm T}$ by the first camera can be calculated by 
\begin{equation} \label{eq:cam1}
[\mathbf{P}(\mathbf{x};1)]=\begin{pmatrix}f_{x}x/z\\
f_{y}y/z\\
1
\end{pmatrix}\equiv \begin{pmatrix}x_{1}\\
y_{1}\\
1
\end{pmatrix},
\end{equation}
where $(\mathbf{x};1)=(x,y,z,1)^{\rm T}$ and $[\cdot]$ denotes the homogeneous coordinate representation. 
The matrix for the second camera is
\begin{equation}
\mathbf{P}'=\mathbf{K}'(\mathbf{I}\,|\,\mathbf{t})=\begin{pmatrix}f_{x} &  0 & c_{x}\\
0 & f_{y} & c_{y}\\
0 & 0 & 1
\end{pmatrix}\begin{pmatrix}1 & 0 & 0 & t_{x}\\
0 & 1 & 0 & t_{t}\\
0 & 0 & 1 & 0
\end{pmatrix},
\end{equation}
and accordingly $\mathbf{x}$ is imaged in the second camera at 
\begin{equation}\label{eq:x_loc}
\begin{pmatrix}x_{2}\\
y_{2}
\end{pmatrix}=\begin{pmatrix}\frac{f_{x}(x+t_{x})}{z}+c_{x}\\
\frac{f_{y}(y+t_{y})}{z}+c_{y}
\end{pmatrix}.
\end{equation}
With \eqref{eq:cam1} and \eqref{eq:x_loc}, coordinates of the correspondence between the two images can be related by
\begin{equation} \label{eq:flow_eq}
\begin{pmatrix}x_{2}\\
y_{2}
\end{pmatrix}-\begin{pmatrix}x_{1}\\
y_{1}
\end{pmatrix}=\begin{pmatrix}\frac{f_{x}t_{x}}{z}+c_{x}\\
\frac{f_{y}t_{y}}{z}+c_{y}
\end{pmatrix}.
\end{equation}
The above equation naturally leads to the formulation of subproblem (2) in the paper, which aims at minimizing the squared difference between the rough flow $W_{\rm rough}$ and the flow converted from depth $W_{\rm convt}$.
Note that since $f_x$, $f_y$ are assumed to be known, in subproblem (2) they are respectively absorbed into $t_x$ and $t_y$ for simplicity.

Notice that \eqref{eq:flow_eq} also plays a crucial role in the data augmentation based on multi-view geometry (Section\,4.3 in the paper). 
With \eqref{eq:flow_eq} we can generate images taken by the second camera given the depth information from the perspective of the first camera.
Specifically, given randomly sampled $\{t_x, t_y, c_x, c_y\}$, the right hand side of \eqref{eq:flow_eq} defines the underlying optical flow from the first image to the second image, which is used to warp the color image into a novel view defined by those parameters. 

\subsection{Detailed Network Architectures}
We used FlowNetC for the cross-modal optical flow estimation; therefore, we refer the readers to \cite{dosovitskiy2015flownet} for its detailed architecture. 
Since FlowNetC takes two three-channel images as inputs, we repeat the one-channel ToF amplitude for three times before feeding it to FlowNetC.
The detailed architectures of the flow fusion network and the backbone of the depth refinement network are provided in Table\,\ref{tab:depth_net}.
Both of these networks are U-Nets with skip connections.
\begin{table}[htbp]
\footnotesize
% \fontsize{7.5pt}{7.5pt}\selectfont
  \centering
  \renewcommand{\arraystretch}{1}
   \begin{tabular}{|m{29pt}<{\centering}|m{8pt}<{\centering}m{3pt}<{\centering}m{27pt}<{\centering}|m{5pt}<{\centering}m{5pt}<{\centering}|m{75pt}<{\centering}|}
    %\begin{tabular}{|c|c|c|c|c|c|c|}
    \hline
   \multicolumn{7}{|c|}{\bf Optical Flow Refinement Network}\bigstrut\\
   \hline
   \textbf{Layer} & \textbf{K} & \textbf{S} & \textbf{Channels} & \textbf{I} & \textbf{O} & \textbf{Input Channels} \bigstrut\\
    \hline
    conv0 & 3$\times$3   & 1     & 4/64    & 1 & 1 & $W_{\rm rough}$, $W_{\rm convt}$ \bigstrut[t]\\
    conv1 & 3$\times$3   & 2     & 64/64   & 1 & 2 & conv0 \\
    conv1\_1 & 3$\times$3   & 1  & 64/128  & 2 & 2 & conv1 \\
    conv2 & 3$\times$3   & 2     & 128/128 & 2 & 4 & conv1\_1 \\
    conv2\_1 & 3$\times$3   & 1  & 128/128 & 4 & 4 & conv2 \bigstrut[b]\\
    $W_{\rm refn}^{(2)}$ & 3$\times$3   & 1     & 128/2 & 4 & 4 & conv2\_1 \bigstrut\\
    upconv1 & 4$\times$4   & 2     & 130/128 & 4 & 2 & conv2\_1, $W_{\rm refn}^{(2)}$ \bigstrut[t]\\
    rconv1 & 3$\times$3   & 1     & 256/64 & 2 & 2 & upconv1, conv1\_1 \bigstrut[b]\\
    $W_{\rm refn}^{(1)}$ & 3$\times$3   & 1     & 64/2  & 2 & 2 & rconv1 \bigstrut\\
    upconv0 & 4$\times$4   & 2     & 66/64 & 2 & 1 & rconv1, $W_{\rm refn}^{(1)}$ \bigstrut[t]\\
    rconv0 & 3$\times$3   & 1     & 128/64 & 1 & 1 & upconv0, conv0 \bigstrut[b]\\
    \hline
    $W_{\rm refn}$ & 3$\times$3   & 1     & 64/2  & 1 & 1 & rconv0 \bigstrut\\
    \hline
    \hline
    \multicolumn{7}{|c|}{\bf Backbone U-Net of ToF-KPN}\bigstrut\\
   \hline
    conv0 & 3$\times$3   & 1        & 5/64    & 1 & 1 & $I_{\rm ToF}\circ W_{\rm refn}$, $D_{\rm ToF} \circ W_{\rm refn}$, $I_{\rm RGB}$ \bigstrut[t]\\
    conv0\_1 & 3$\times$3   & 1     & 64/64   & 1 & 1 & conv0 \\
    conv1 & 3$\times$3   & 2        & 64/128  & 1 & 2 & conv0\_1 \\
    conv1\_1 & 3$\times$3   & 1     & 128/128 & 2 & 2 & conv1 \\
    conv2   & 3$\times$3   & 2      & 128/128 & 2 & 4 & conv1\_1  \bigstrut[b]\\
    conv2\_1 & 3$\times$3   & 1     & 128/128 & 4 & 4 & conv2 \bigstrut[t]\\
    conv3    & 3$\times$3   & 2     & 128/256 & 8 & 8 & conv2\_1 \bigstrut[b]\\
    conv3\_1    & 3$\times$3   & 1  & 256/256 & 8 & 8 & conv3\bigstrut[b]\\
    upconv0 & 3$\times$3   & 2      & 256/128 & 8 & 4 & conv3\_1 \bigstrut\\
    upconv0\_1 & 4$\times$4   & 1   & 128/128 & 4 & 4 & upconv0 \bigstrut[t]\\
    upconv1 & 3$\times$3   & 2      & 256/128 & 4 & 2 & conv2\_1, upconv0\_1 \bigstrut\\
    upconv1\_1 & 4$\times$4   & 1   & 128/128 & 2 & 2 & upconv1 \bigstrut[t]\\
    upconv2 & 3$\times$3   & 2      & 256/64  & 2 & 1 & upconv1, conv1\_1 \bigstrut\\
    upconv2\_1 & 4$\times$4   & 1   & 64/64   & 1 & 1 & upconv2 \bigstrut[b]\\
    \hline
    ${\bf w}$, $b$ & 3$\times$3   & 1     & 64/10   & 1 & 1 & upconv2 \bigstrut\\
    \hline
    \end{tabular}%
    \vspace{3pt}
    \caption{Network architecture of the optical flow fusion network and the backbone of our ToF-KPN. $I_{\rm ToF}\circ W_{\rm refn}$ denotes the ToF amplitude image warped by the flow $W_{\rm refn}$ and $D_{\rm ToF} \circ W_{\rm refn}$ similarly denotes the warped ToF depth image. {\bf K} means kernel size, {\bf S} means stride, and {\bf Channels} is the number of input and output channels. {\bf I} and {\bf O} are the input and output downsampling factor relative to the input. Separation by ``," in the {\bf Input Channels} means concatenation.}
  \label{tab:depth_net}%
\end{table}%

\section{More Details on Data Generation and Pre-processing}\label{sec:data}

This section briefly reviews the background of synthetic data generation and explains how to get simulated ToF depth from transient rendering. 
We also describe the data pre-processing procedure being used in our work.

\subsection{More on Synthetic Data Generation}
Transient rendering \cite{jarabo2014framework,smith2008transient} is a tool from computer graphics used to study the propagation of light in extremely short timescales.
For ToF sensor with a single light source, transient rendering can be regarded as simulating the \emph{temporal point spread function} (TPSF) of each pixel in the image that depends both on the camera and the scene. 
A TPSF encodes the temporal energy distribution of the homecoming light at its pixel.
In case there is no MPI, the TPSF will be an impulse peaking at the true depth, otherwise the TPSF will have a scene-dependent tail.
During rendering, we also adopt the assumption that the scenes contain mainly diffusive materials \cite{su2018deep,gupta2015phasor, kadambi2017rethinking}, which is valid for most real-life scenarios.
% Note that this assumption also applies to the statement that GHz frequencies has little MPI effect \cite{gupta2015phasor, kadambi2017rethinking}, since it requires the non-peak part of TPSF to be essentially constant at GHz scale. 
Then each pixel of the raw ToF signal can be modeled as the integral over the exposure time of the temporal convolution between the modulated light and the TPSF. 
Since the TPSF captures the multi-path interference, it faithfully approximates the errors of ToF sensors in real life. 
We refer the readers to \cite{heide2013low} for more mathematical details in this respect.

\begin{figure*}[t]
\vspace{-20pt}
\centering \scriptsize
\newlength{\myheight}
\setlength{\myheight}{60pt}
    \captionsetup[subfigure]{labelformat=empty}
    \subfloat{\includegraphics[height=\myheight]{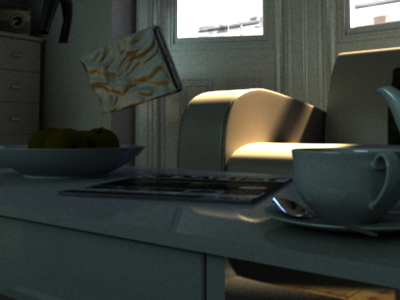}}\hspace{\inter}
    \subfloat{\includegraphics[height=\myheight]{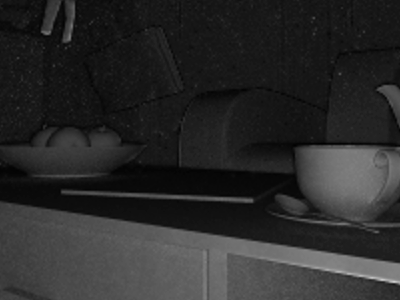}} \hspace{\inter}
    \subfloat {\includegraphics[height=\myheight]{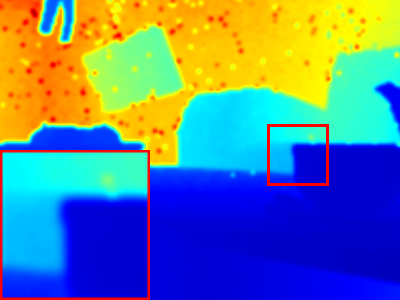}}\hspace{\inter}
    \subfloat{\includegraphics[height=\myheight]{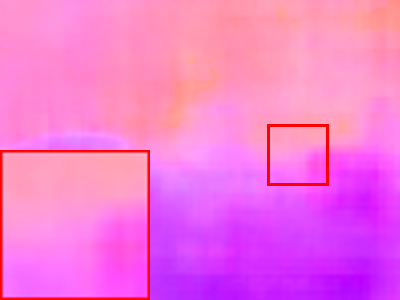}}\hspace{\inter}
    \subfloat{\includegraphics[height=\myheight]{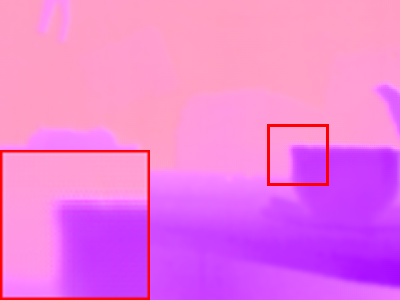}}\hspace{\inter}
    \subfloat{\includegraphics[height=\myheight]{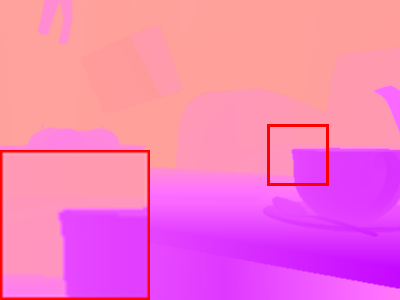}} \\[-3ex]
     \subfloat{\includegraphics[height=\myheight]{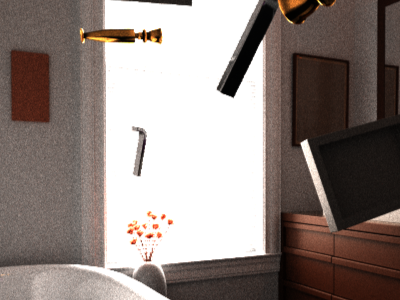}}\hspace{\inter}
    \subfloat{\includegraphics[height=\myheight]{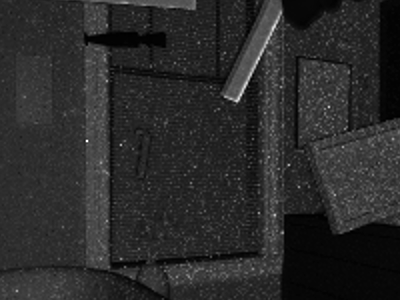}} \hspace{\inter}
    \subfloat {\includegraphics[height=\myheight]{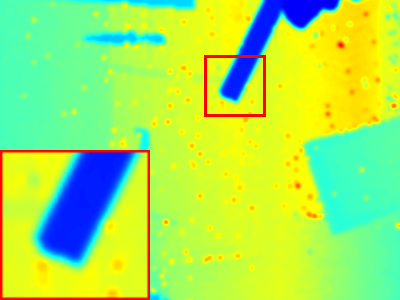}}\hspace{\inter}
    \subfloat{\includegraphics[height=\myheight]{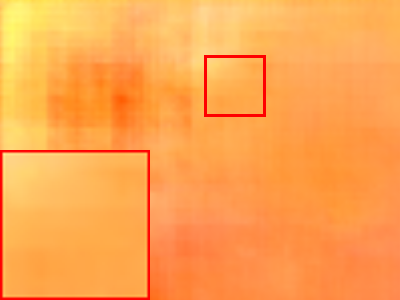}}\hspace{\inter}
    \subfloat{\includegraphics[height=\myheight]{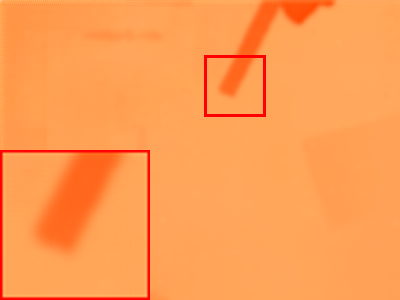}}\hspace{\inter}
    \subfloat{\includegraphics[height=\myheight]{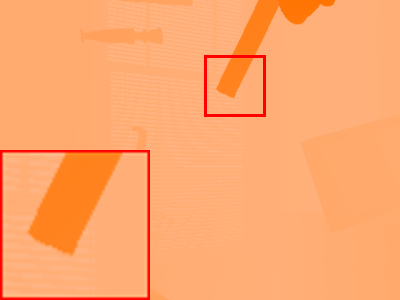}} \\[-3ex]
     \subfloat[RGB image]{\includegraphics[height=\myheight]{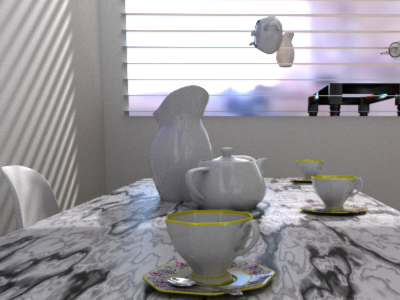}}\hspace{\inter}
    \subfloat[ToF amplitude]{\includegraphics[height=\myheight]{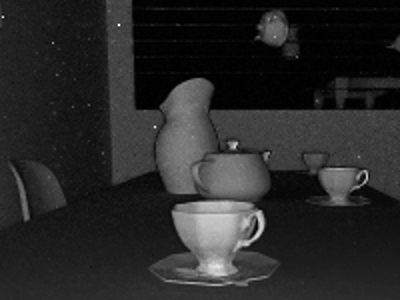}} \hspace{\inter}
    \subfloat[ToF depth image]{\includegraphics[height=\myheight]{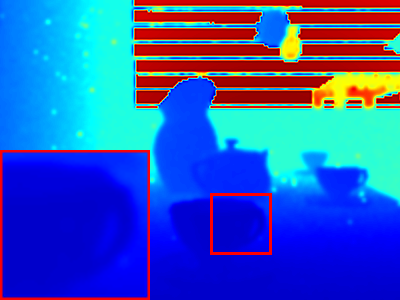}}\hspace{\inter}
    \subfloat[$W_{\rm rough}$]{\includegraphics[height=\myheight]{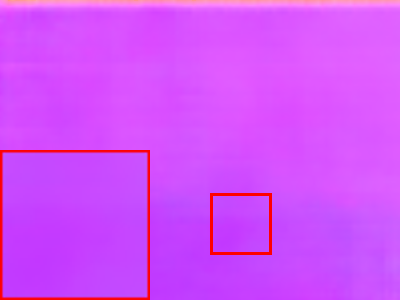}}\hspace{\inter}
    \subfloat[$W_{\rm refn}$]{\includegraphics[height=\myheight]{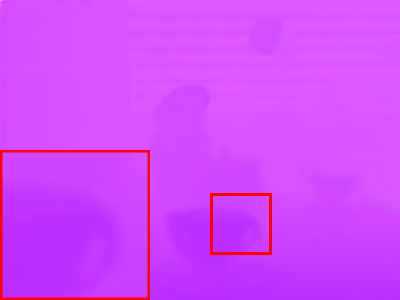}}\hspace{\inter}
    \subfloat[$W_{\rm gt}$]{\includegraphics[height=\myheight]{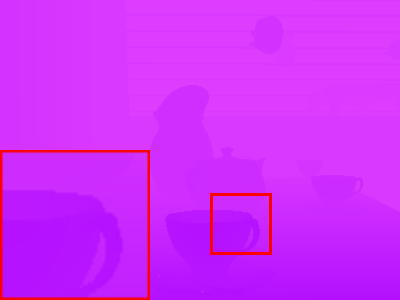}}
    \caption{Visual comparisons before and after optical flow refinement. The refinement incorporates ToF depth image via a depth-to-flow conversion, which greatly enhances the accuracy of cross-modal optical flow estimation.}
 \label{fig:supp_flow_refine}
\end{figure*}

To generate the synthetic ToF measurements, let $\{I_{t}\}_{t=1}^{T}$ be the transient images of a scene under the point light source of the ToF sensor.
We also let $L_{\rm sin}^{(\omega)}$, $L_{\rm cos}^{(\omega)}$ be the sine and cosine light waves with frequency $\omega$, respectively. 
Then, the ToF correlation images at pixel $p$ are obtained by:
\begin{equation}
\begin{split}
C_{\rm sin}(p, \omega) &= \sum\nolimits_{t=1}^{T} I_t(p) \cdot L_{\rm sin}^{(\omega)}(t),\\
C_{\rm cos}(p, \omega) &= \sum\nolimits_{t=1}^{T} I_t(p) \cdot L_{\rm cos}^{(\omega)}(t),
\end{split}
\end{equation}
where $\{I_t(p)\}_{t=1}^T$ is simply the TPSF at the pixel $p$.
The phase angle at pixel $p$ used for depth conversion can then be determined by, {\it e.g.}, taking the argument of the complex number $C_{\rm cos}(p, \omega) + iC_{\rm sin}(p, \omega)$.

Furthermore, note that the depth obtained above in fact measures the distance from the scene point to the {\it light source}, rather than to the {\it image plane}, where the latter is used in our work (recall Equation \eqref{eq:cam1} in Section\,\ref{ssec:derivation}).
Therefore, in our synthetic dataset we also perform standard plane correction \cite{hartley2003multiple} to the depth obtained above so as to convert point-to-point distance to point-to-plane distance.

% Note that adding noise to the transient images and the correlation images is also a more faithful way to approximate environmental and sensor noises.

%Finally, we would like to acknowledge the authors of the beautiful Blender scenes\footnote{https://www.blendswap.com/blends/view/73937}\footnote{https://www.blendswap.com/blends/view/75302}\footnote{https://www.blendernation.com/2013/03/27/model-download-barcelona-pavilion}\footnote{https://www.blendswap.com/blends/view/41683}\footnote{https://www.blendswap.com/blends/view/75431} utilized in the generation of our synthetic dataset.

\subsection{Data Pre-processing}
%\paragraph{A remark on pre-processing ToF amplitude images.}

Since raw ToF amplitude images, ToF depth images and the captured RGB images initially have intensities of different scales, proper data pre-processing and normalization is helpful for the training of neural networks \cite{ioffe2015batch}, especially when the Siamese network ({\it i.e.}, the FlowNetC) is used in our work \cite{dosovitskiy2015flownet}.
We first present our pre-processing procedure for the ToF amplitude images.
The ToF amplitude images often exhibit extremely high contrast between the foreground and the background of the captured scenes.
Simply re-scaling a ToF amplitude image into the range $[0,1]$ ({\it i.e.}, divide the image by its maximum intensity), or truncation ({\it i.e.}, set all values above certain threshold to be a same value) may result in overly dark or overly bright regions.
Such an unbalanced intensity difference brings negative impact for the rough flow estimation as well as depth refinement. 
We have thus adopted a simple pre-processing to the raw ToF amplitude images to mitigate such effects. 

Our approach is based on the fact that, the intensity of the ToF amplitude obeys an inverse square relationship to the distance. 
Specifically, we found the following simple pixel-wise transform to work well in practice:
\begin{equation}
% I_{\rm ToF} = \frac{I_{\rm ToF}^{\rm (raw)} \odot D_{\rm ToF}^2}{{\rm mean}(I_{\rm ToF}^{\rm (raw)} \odot D_{\rm ToF}^2)},
I_{\rm ToF} = I_{\rm ToF}^{\rm (raw)} \odot D_{\rm ToF}^2,
\end{equation}
where $\odot$ denotes the \emph{Hadamard product}, $D_{\rm ToF}^2$ denotes the pixel-wise squaring of the ToF depth image $D_{\rm ToF}$,
%${\rm mean}(\cdot)$ denotes the operation of the taking average value
 and $I_{\rm ToF}^{\rm (raw)}$ denotes the raw capture of ToF amplitude. 
The above normalization scheme can not only mitigate the huge contrast difference, but also make the overall brightness of different scenes roughly similar.
After that, the obtained ToF amplitude images are further normalized to the range $[0,1]$.
All of the ToF amplitude images in our work, synthetic or real, used in training or testing, have been pre-processed with the above normalization scheme.

Data normalization is also performed for the RGB images and the depth images (both the ToF depth images and the ground-truth depth images).
Specifically, for both of the training and testing, the RGB images and the depth images are all normalized to the interval $[0,1]$.

\section{More Experimental Results}\label{sec:results}

\begin{figure*}[t]
% \vspace{-20pt}
\centering \scriptsize
% \newlength{\myheight}
\setlength{\myheight}{72pt}
    \captionsetup[subfigure]{labelformat=empty}
    \subfloat{\includegraphics[height=\myheight]{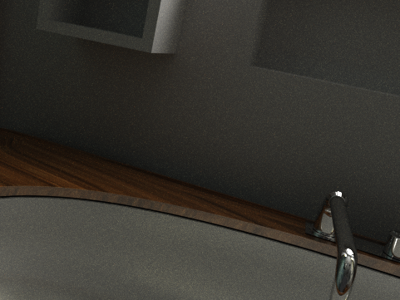}}\hspace{\inter}
    \subfloat{\includegraphics[height=\myheight]{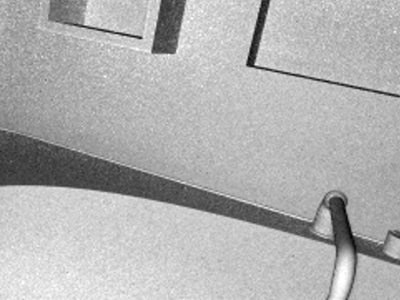}} \hspace{\inter}
    \subfloat {\includegraphics[height=\myheight]{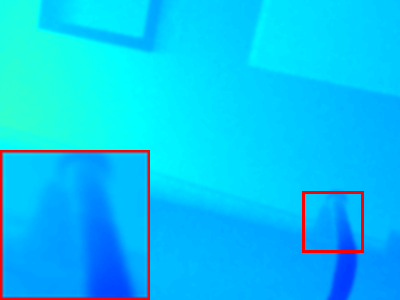}}\hspace{\inter}
    \subfloat{\includegraphics[height=\myheight]{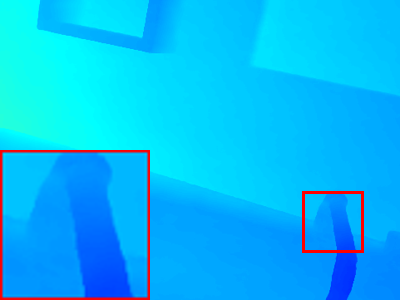}}\hspace{\inter}
    \subfloat{\includegraphics[height=\myheight]{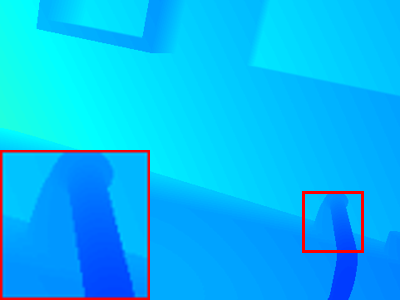}}\hspace{\inter} \\[-3ex]
    
    \subfloat{\includegraphics[height=\myheight]{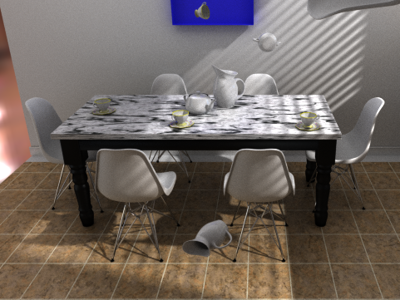}}\hspace{\inter}
    \subfloat{\includegraphics[height=\myheight]{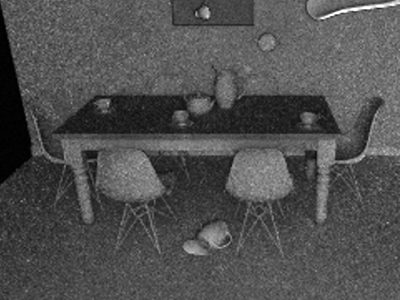}} \hspace{\inter}
    \subfloat {\includegraphics[height=\myheight]{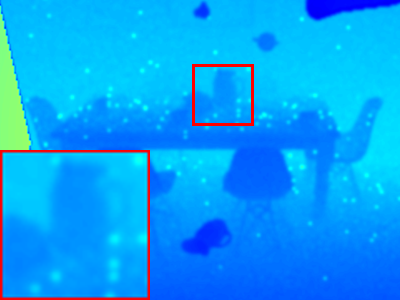}}\hspace{\inter}
    \subfloat{\includegraphics[height=\myheight]{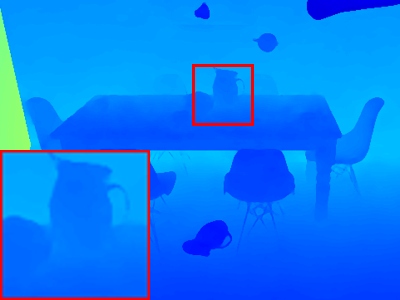}}\hspace{\inter}
    \subfloat{\includegraphics[height=\myheight]{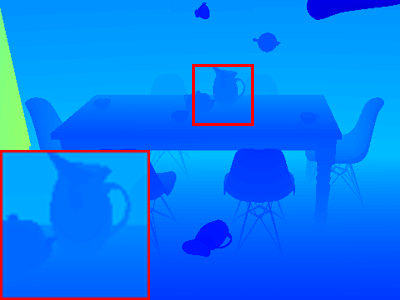}}\hspace{\inter} \\[-3ex]
    
    \subfloat[RGB image]{\includegraphics[height=\myheight]{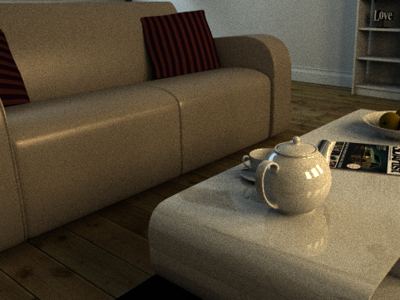}}\hspace{\inter}
    \subfloat[ToF amplitude]{\includegraphics[height=\myheight]{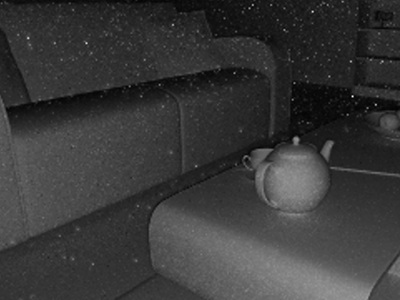}} \hspace{\inter}
    \subfloat[ToF depth imasge]{\includegraphics[height=\myheight]{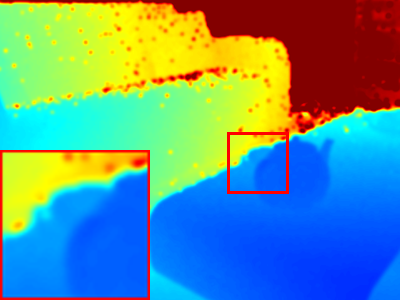}}\hspace{\inter}
    \subfloat[\textsc{ToF-KPN} (ours)]{\includegraphics[height=\myheight]{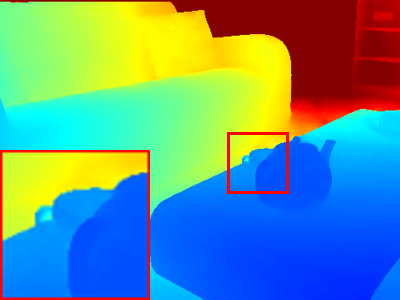}}\hspace{\inter}
    \subfloat[Ground-truth]{\includegraphics[height=\myheight]{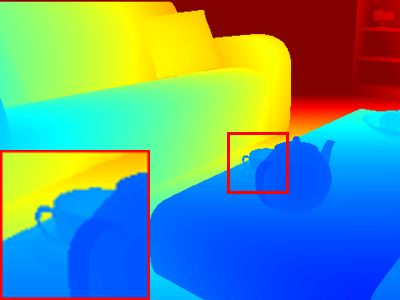}}\hspace{\inter} 
 \caption{Visual results of the ToF-KPN module for depth image refinement on the ToF-FlyingThings3D dataset.}
 \label{fig:supp_depth_refine}
\end{figure*}

\begin{figure*}[t]
% \vspace{-20pt}
\centering \scriptsize
% \newlength{\myheight}
\setlength{\myheight}{72pt}
    \captionsetup[subfigure]{labelformat=empty}
    \subfloat{\includegraphics[height=\myheight]{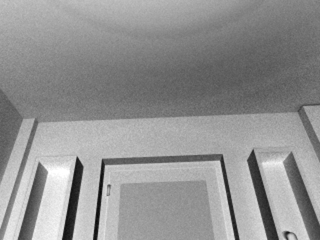}}\hspace{\inter}
    \subfloat{\includegraphics[height=\myheight]{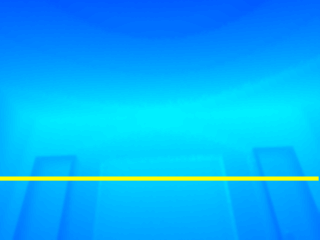}} \hspace{8pt}
    \subfloat {\includegraphics[height=\myheight]{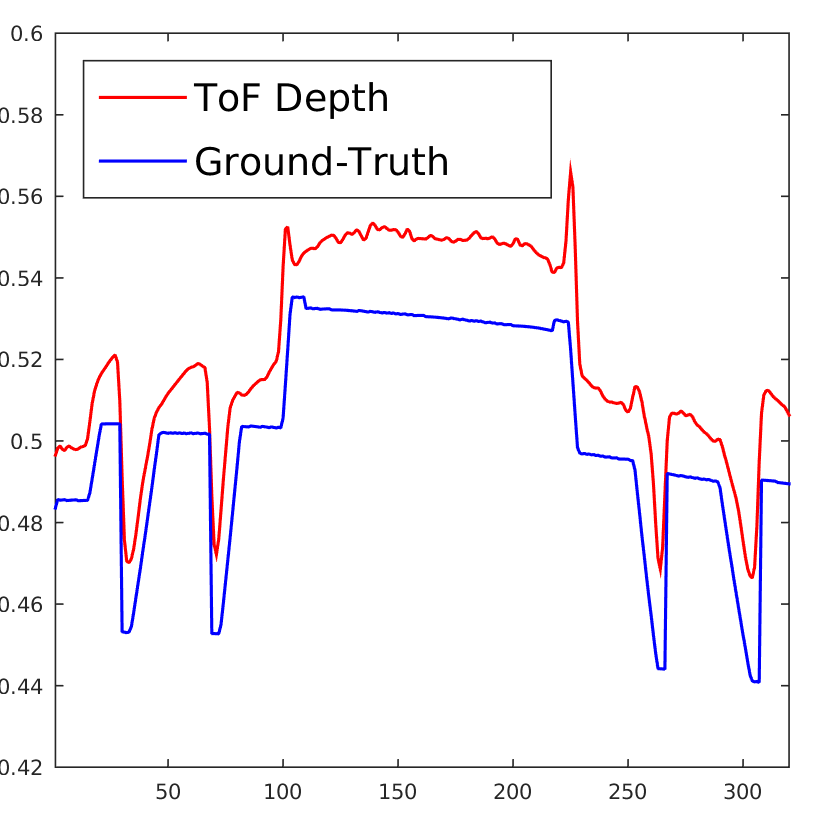}}\hspace{-1pt}
    \subfloat{\includegraphics[height=\myheight]{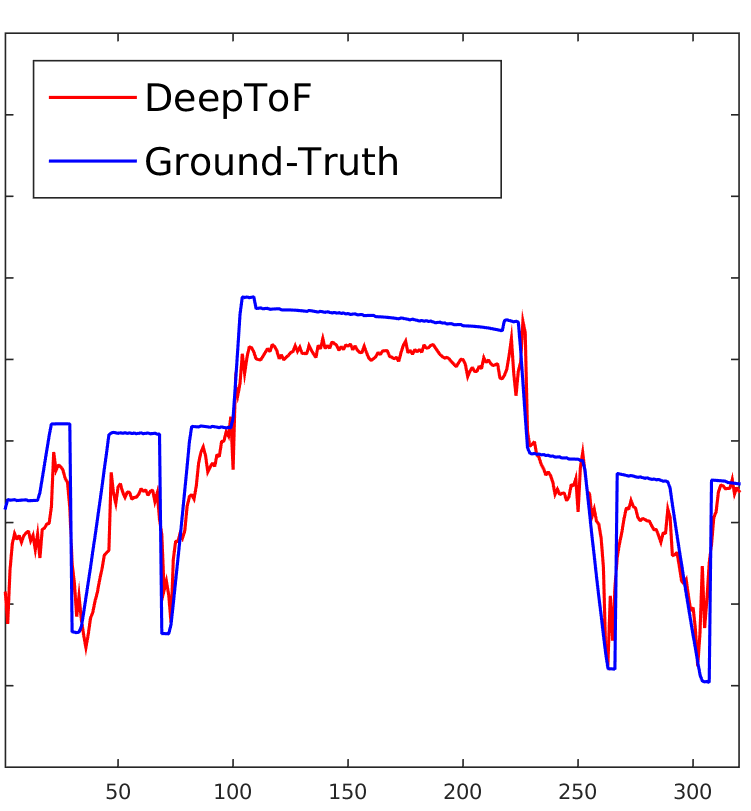}}\hspace{-1pt}
    \subfloat{\includegraphics[height=\myheight]{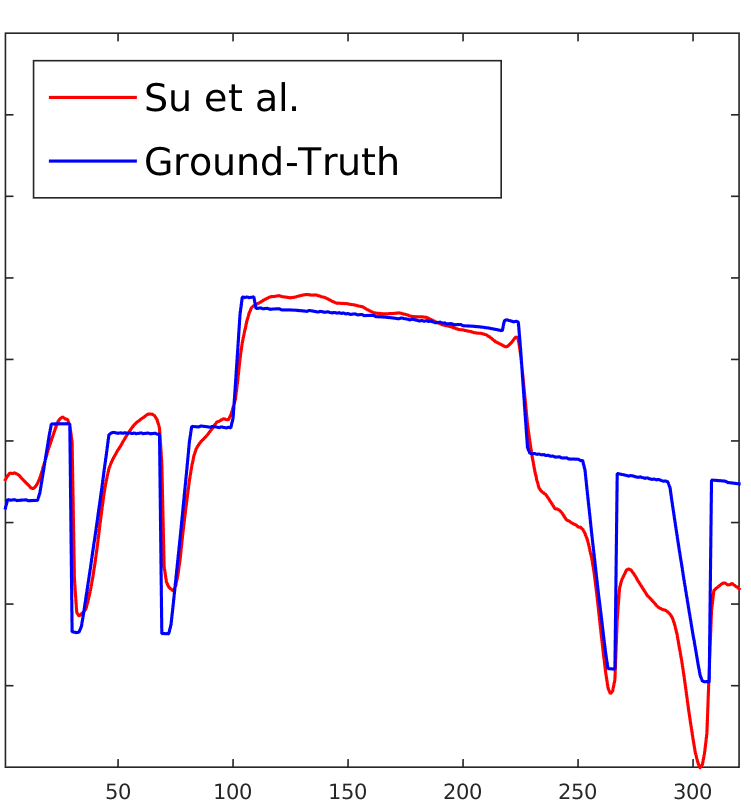}}\hspace{-1pt}
    \subfloat{\includegraphics[height=\myheight]{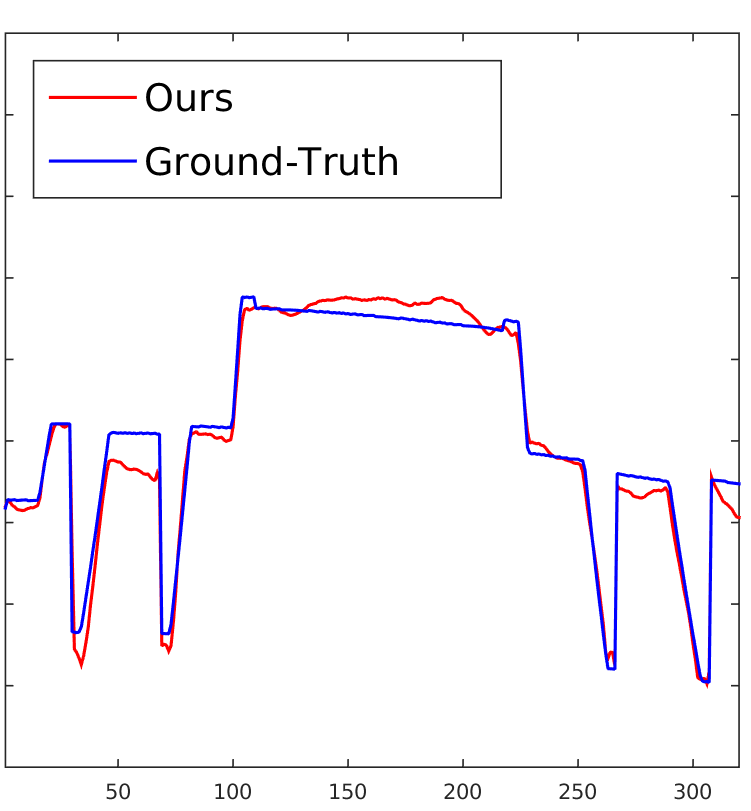}} \\[-3ex]
    \subfloat{\includegraphics[height=\myheight]{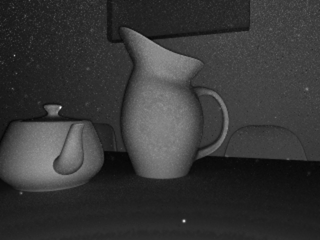}}\hspace{\inter}
    \subfloat{\includegraphics[height=\myheight]{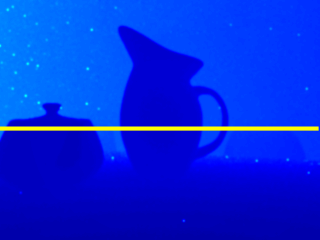}} \hspace{8pt}
    \subfloat {\includegraphics[height=\myheight]{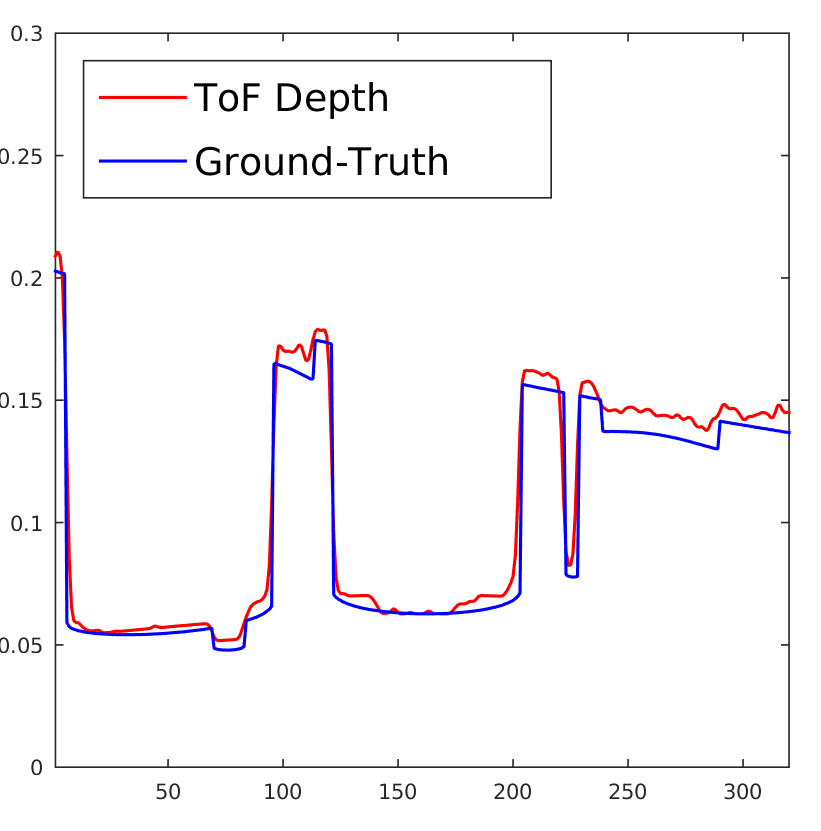}}\hspace{-1pt}
    \subfloat{\includegraphics[height=\myheight]{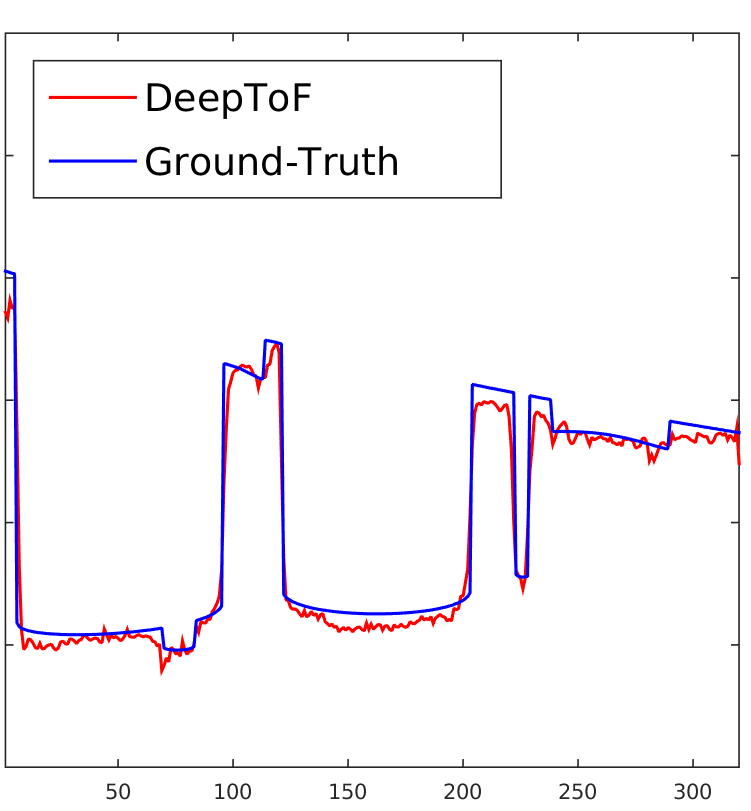}}\hspace{-1pt}
    \subfloat{\includegraphics[height=\myheight]{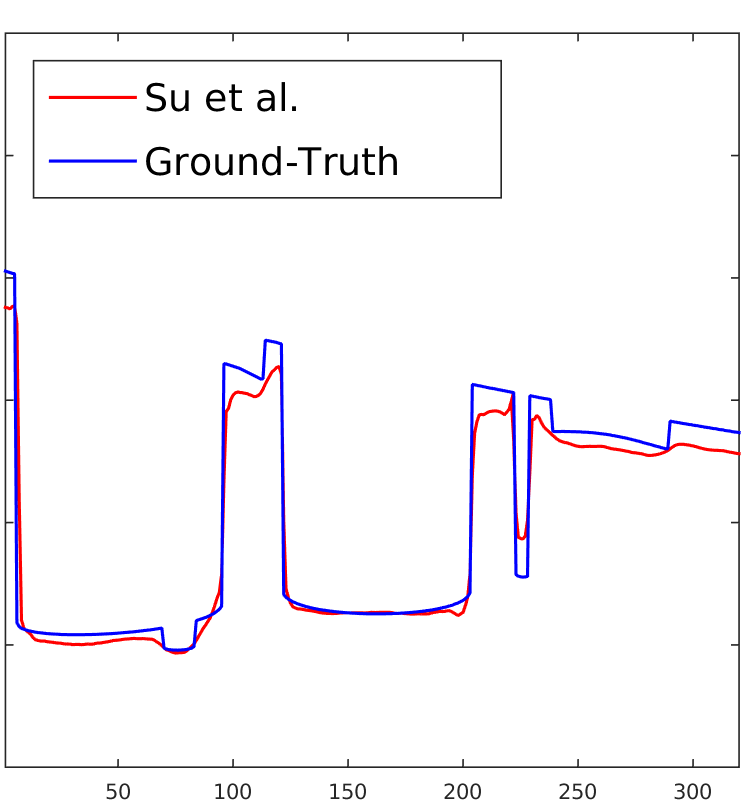}}\hspace{-1pt}
    \subfloat{\includegraphics[height=\myheight]{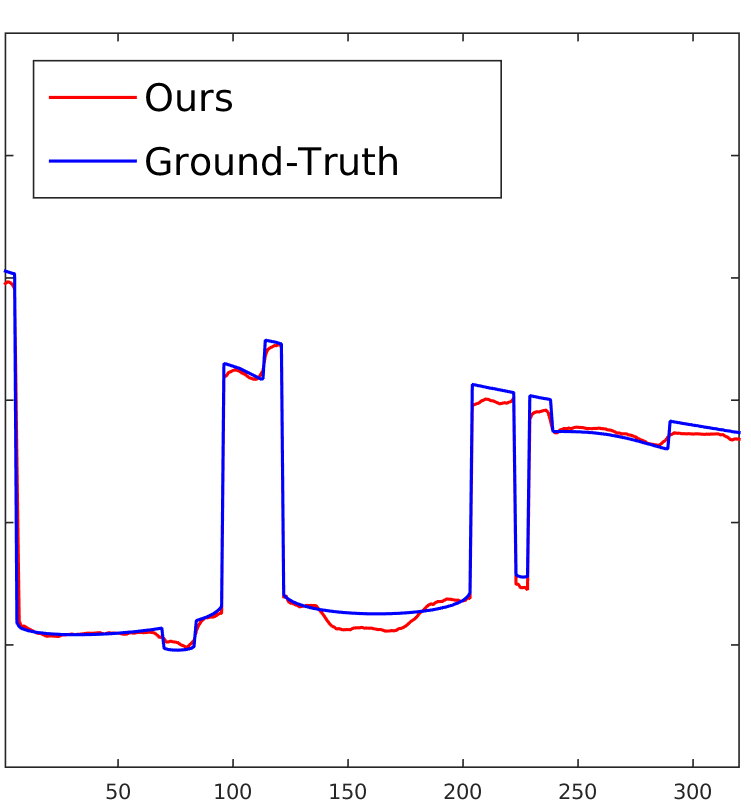}} \\[-3ex]
    \subfloat{\includegraphics[height=\myheight]{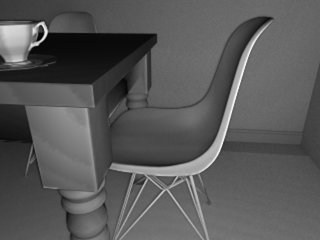}}\hspace{\inter}
    \subfloat{\includegraphics[height=\myheight]{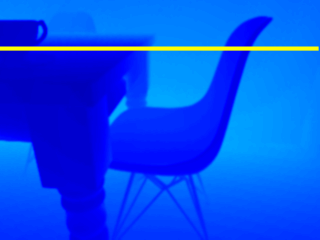}} \hspace{8pt}
    \subfloat {\includegraphics[height=\myheight]{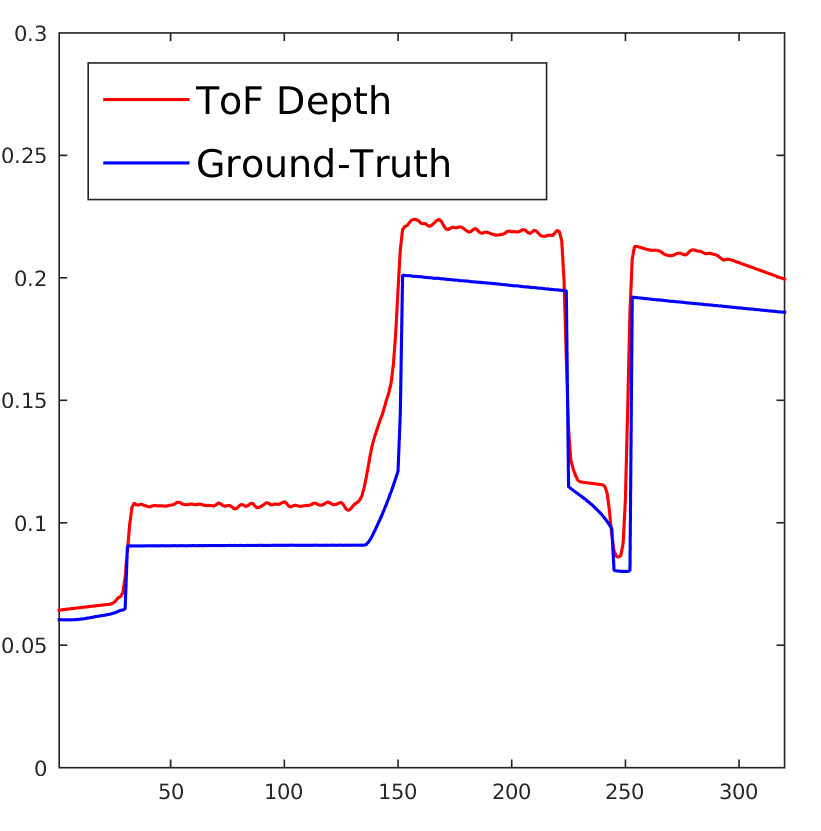}}\hspace{-1pt}
    \subfloat{\includegraphics[height=\myheight]{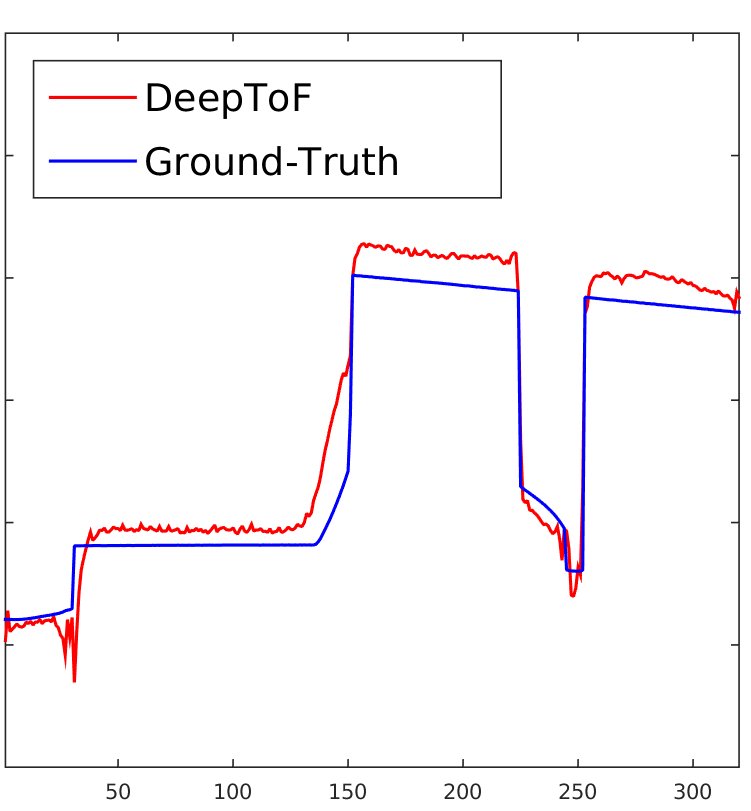}}\hspace{-1pt}
    \subfloat{\includegraphics[height=\myheight]{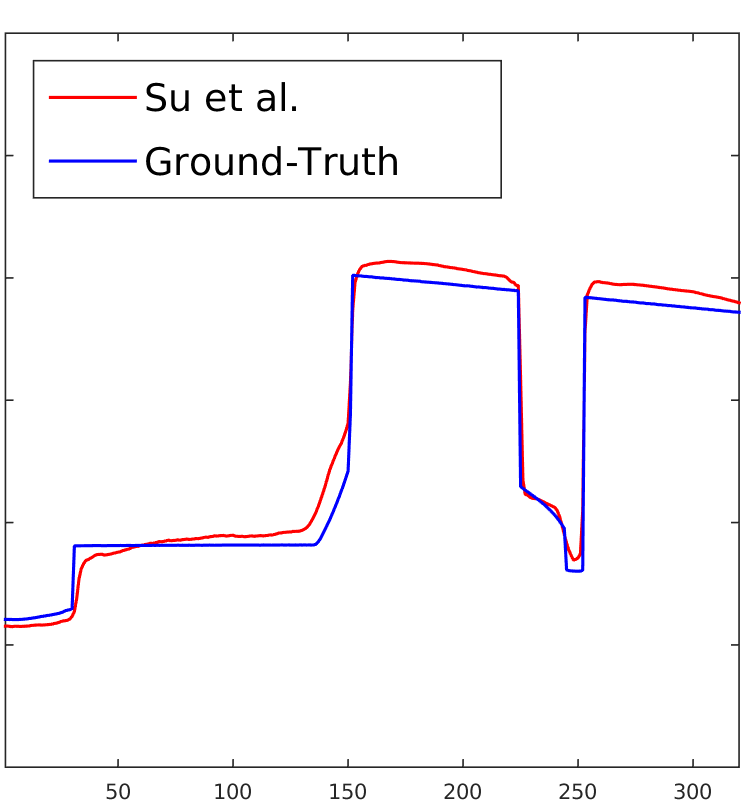}}\hspace{-1pt}
    \subfloat{\includegraphics[height=\myheight]{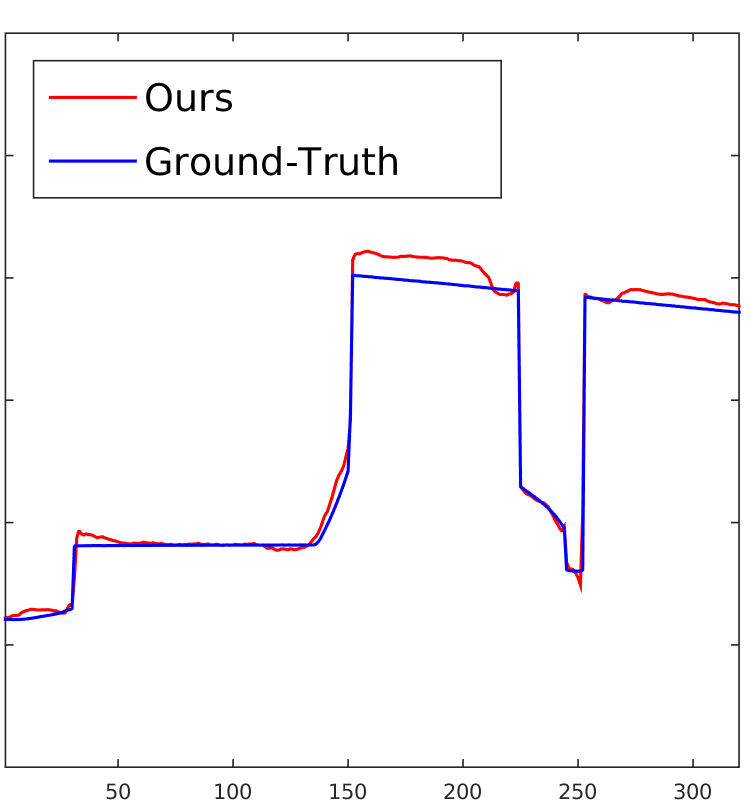}} \\[-3ex]
    \subfloat[ToF amplitude]{\includegraphics[height=\myheight]{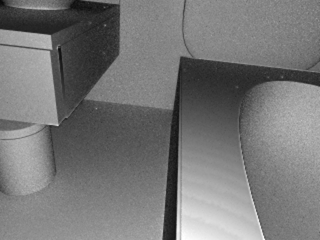}}\hspace{\inter}
    \subfloat[ToF depth image]{\includegraphics[height=\myheight]{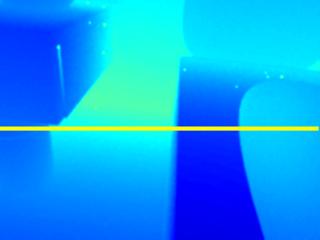}} \hspace{8pt}
    \subfloat[ToF depth values] {\includegraphics[height=\myheight]{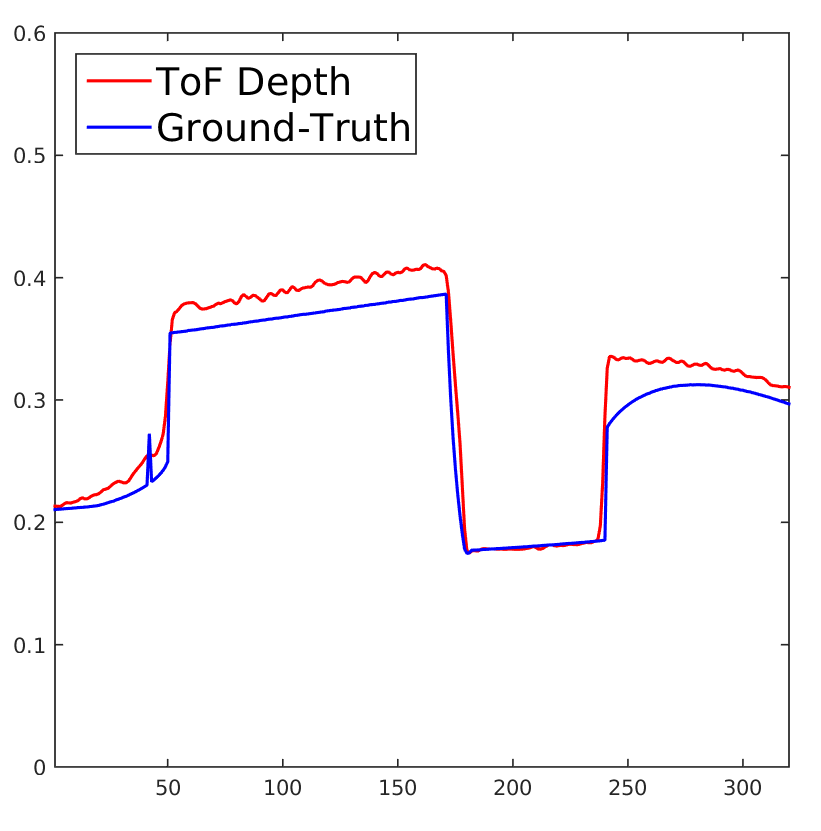}}\hspace{-1pt}
    \subfloat[\textsc{DeepToF}~\cite{marco2017deeptof}] {\includegraphics[height=\myheight]{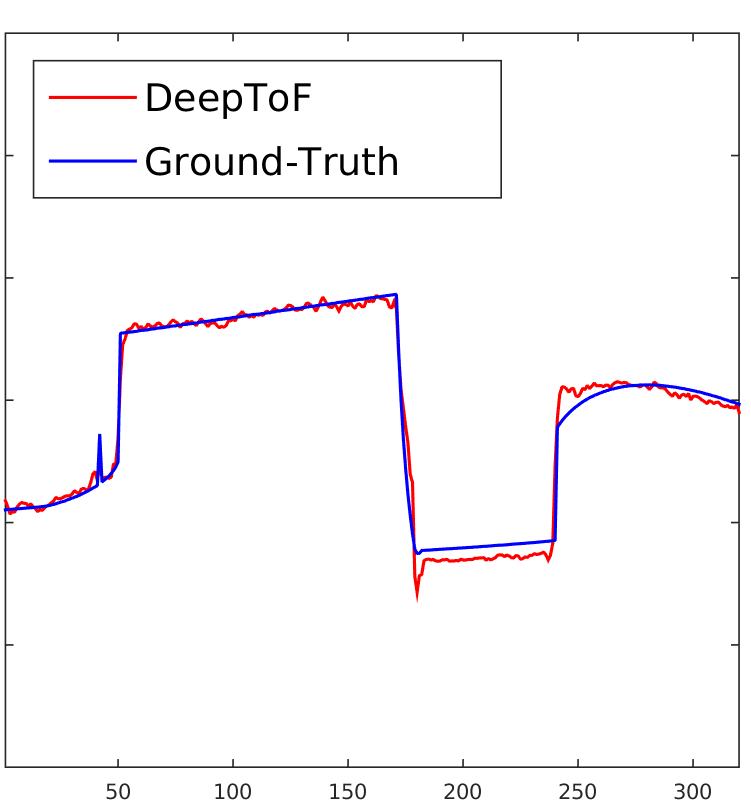}}\hspace{-1pt}
    \subfloat[Su~{\it{et al.}}~\cite{su2018deep}] {\includegraphics[height=\myheight]{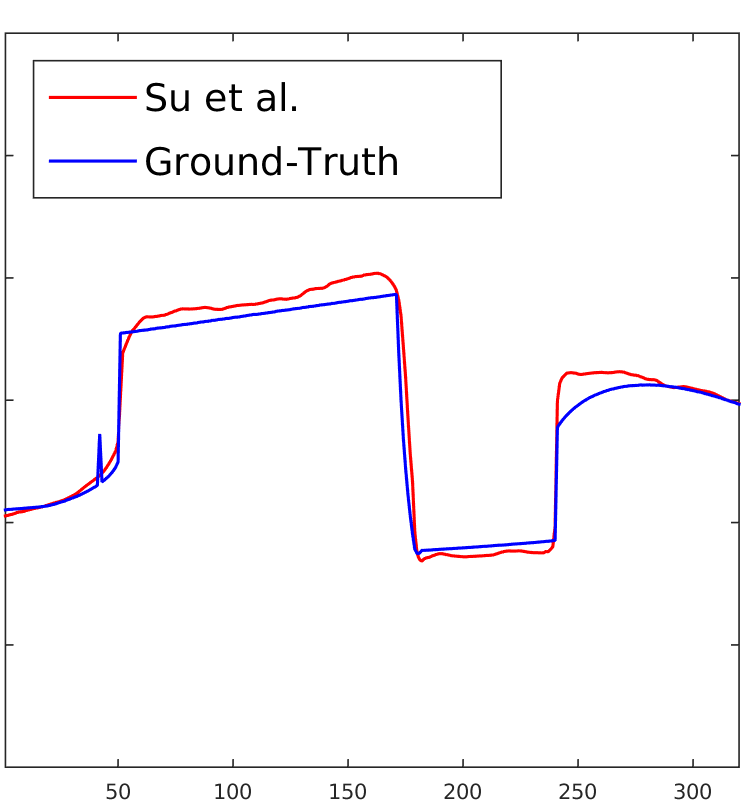}}\hspace{-1pt}
    \subfloat[\textsc{ToF-KPN} (ours)]{\includegraphics[height=\myheight]{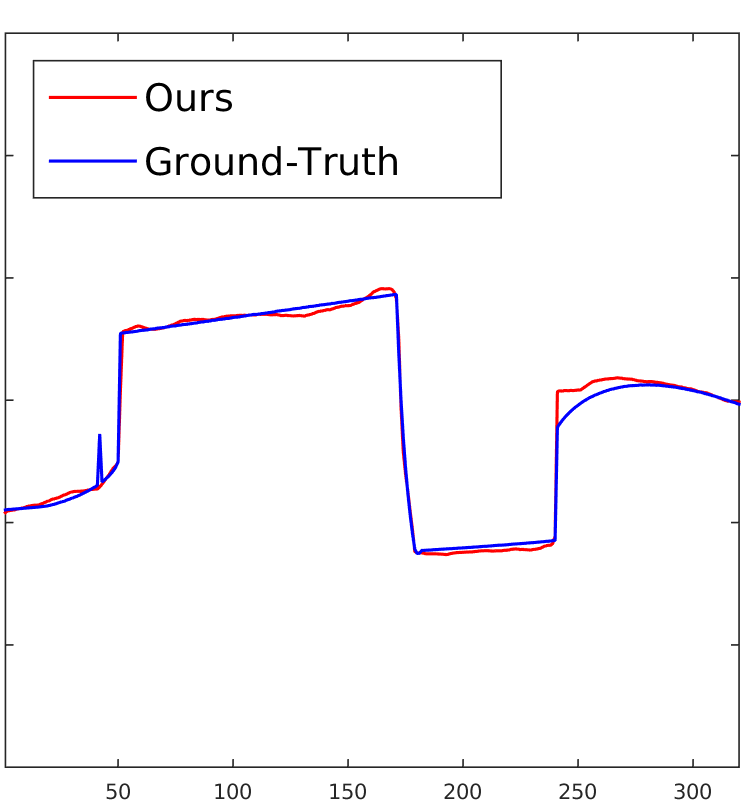}}
 \caption{Depth values of different approaches on a scan-line are shown, alongside with the ground-truth. Note that no color images are used in this experiment.}
 \label{fig:supp_exp_tof}
\end{figure*}

We provide more experimental results in this section.
We first present more results on our optical flow refinement via the ToF depth images.
We then showcase more results demonstrating the effectiveness of our ToF-KPN. 
Finally, more results on our overall DEAR framework are presented.

\begin{figure*}[t]
\newlength{\pipewidth}
\setlength{\pipewidth}{79pt}
\centering \scriptsize
    \subfloat{\includegraphics[width=\pipewidth]{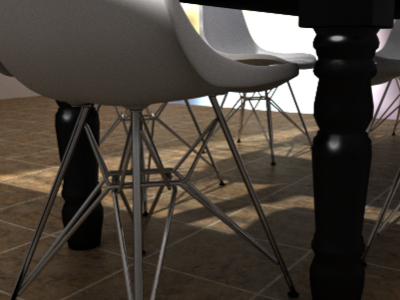}}\hspace{\inter}
    \subfloat{\includegraphics[width=\pipewidth]{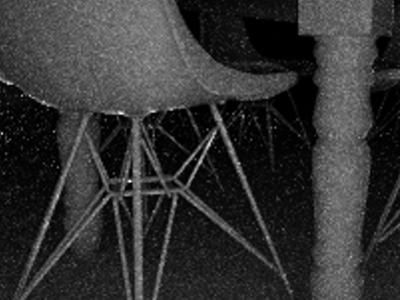}}\hspace{\inter}
    \subfloat{\includegraphics[width=\pipewidth]{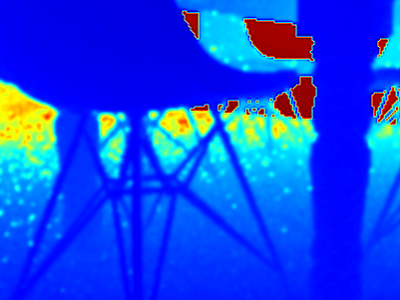}}\hspace{\inter}
    \subfloat{\includegraphics[width=\pipewidth]{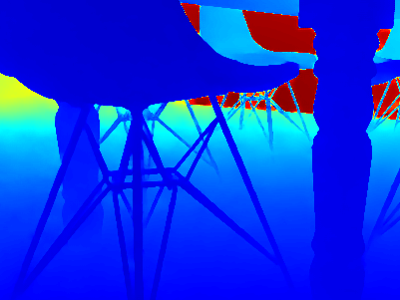}}\hspace{\inter}
    \subfloat{\includegraphics[width=\pipewidth]{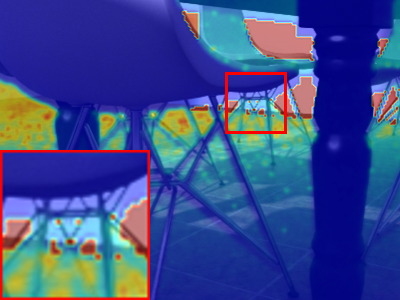}}\hspace{\inter}
    \subfloat{\includegraphics[width=\pipewidth]{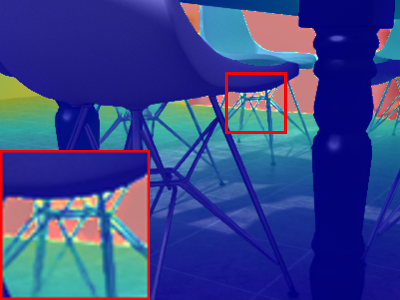}} \\[-3ex]
    \subfloat{\includegraphics[width=\pipewidth]{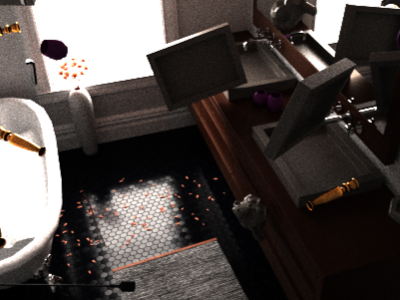}}\hspace{\inter}
    \subfloat{\includegraphics[width=\pipewidth]{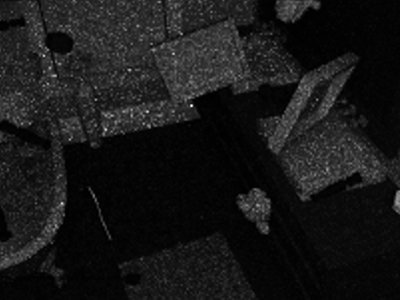}}\hspace{\inter}
    \subfloat{\includegraphics[width=\pipewidth]{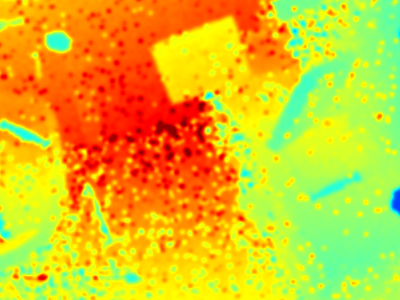}}\hspace{\inter}
    \subfloat{\includegraphics[width=\pipewidth]{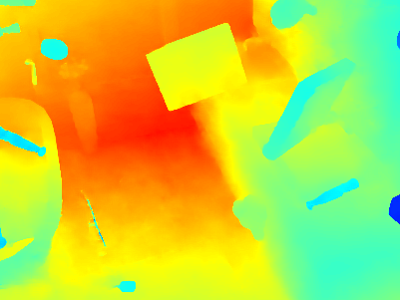}}\hspace{\inter}
    \subfloat{\includegraphics[width=\pipewidth]{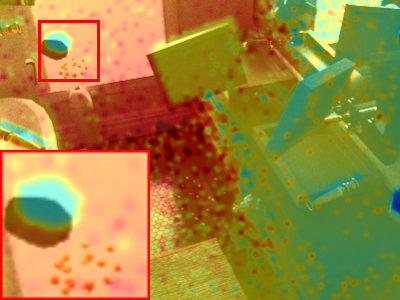}}\hspace{\inter}
    \subfloat{\includegraphics[width=\pipewidth]{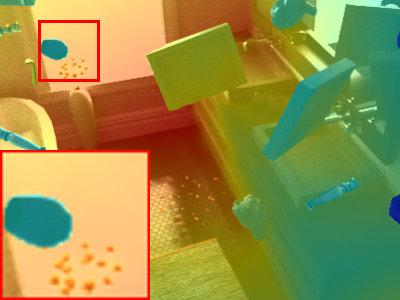}} \\[-3ex]
    \subfloat{\includegraphics[width=\pipewidth]{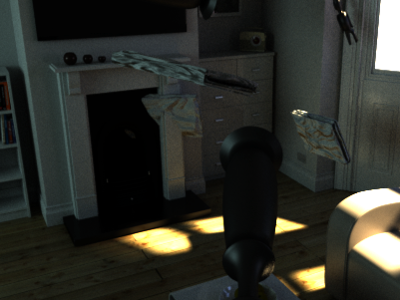}}\hspace{\inter}
    \subfloat{\includegraphics[width=\pipewidth]{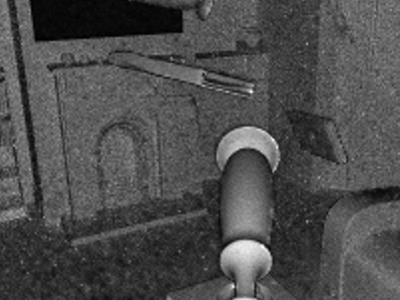}}\hspace{\inter}
    \subfloat{\includegraphics[width=\pipewidth]{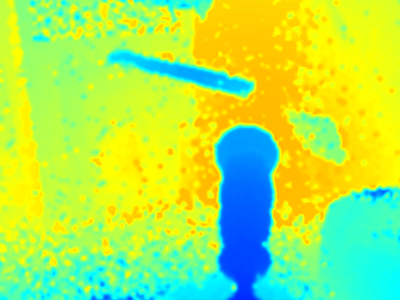}}\hspace{\inter}
    \subfloat{\includegraphics[width=\pipewidth]{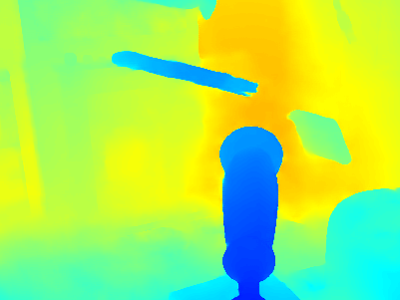}}\hspace{\inter}
    \subfloat{\includegraphics[width=\pipewidth]{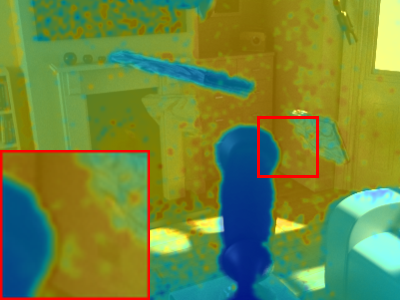}}\hspace{\inter}
    \subfloat{\includegraphics[width=\pipewidth]{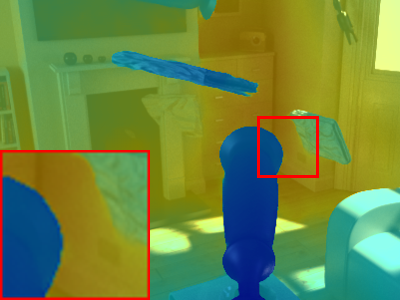}} \\[-3ex]
    \subfloat{\includegraphics[width=\pipewidth]{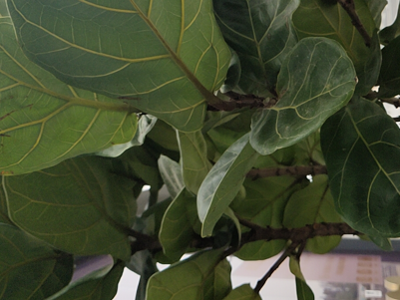}}\hspace{\inter}
    \subfloat{\includegraphics[width=\pipewidth]{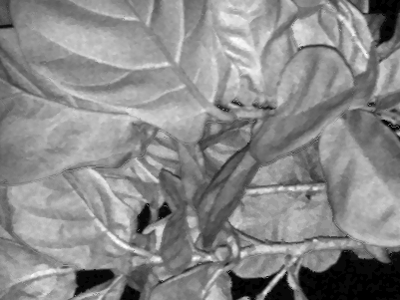}}\hspace{\inter}
    \subfloat{\includegraphics[width=\pipewidth]{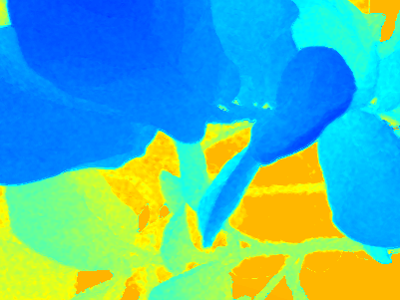}}\hspace{\inter}
    \subfloat{\includegraphics[width=\pipewidth]{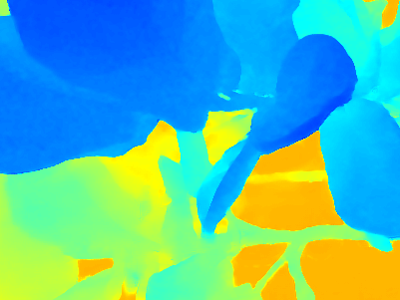}}\hspace{\inter}
    \subfloat{\includegraphics[width=\pipewidth]{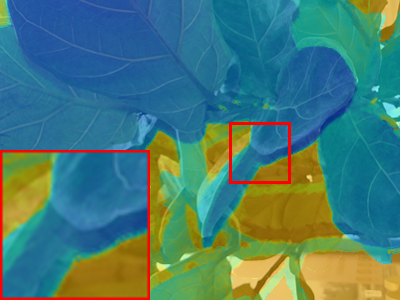}}\hspace{\inter}
    \subfloat{\includegraphics[width=\pipewidth]{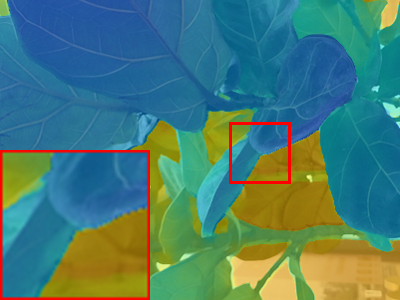}} \\[-3ex]
    \subfloat{\includegraphics[width=\pipewidth]{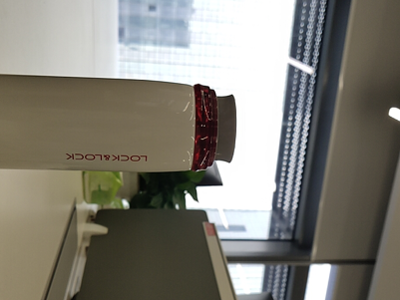}}\hspace{\inter}
    \subfloat{\includegraphics[width=\pipewidth]{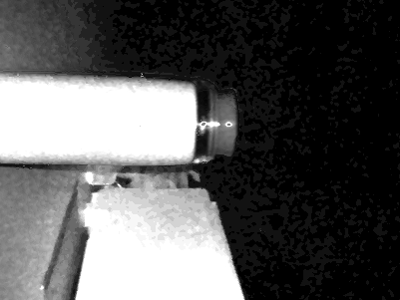}}\hspace{\inter}
    \subfloat{\includegraphics[width=\pipewidth]{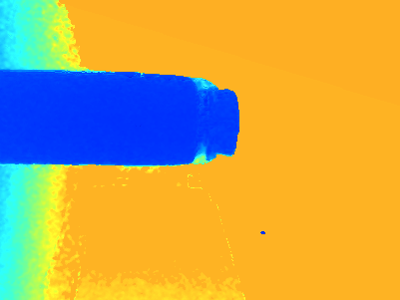}}\hspace{\inter}
    \subfloat{\includegraphics[width=\pipewidth]{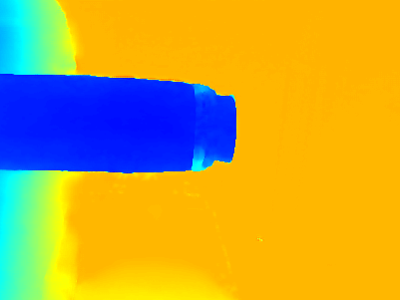}}\hspace{\inter}
    \subfloat{\includegraphics[width=\pipewidth]{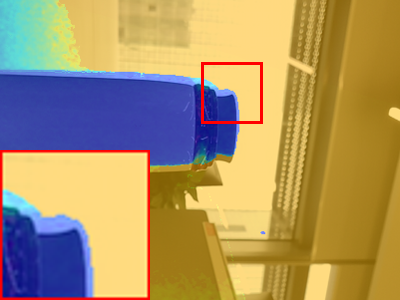}}\hspace{\inter}
    \subfloat{\includegraphics[width=\pipewidth]{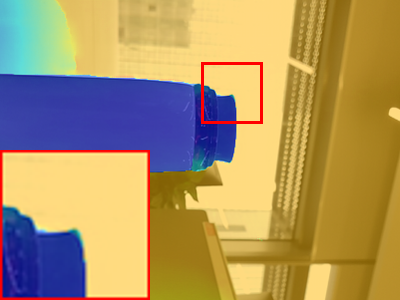}} \\[-3ex]
    \addtocounter{subfigure}{-30}
    \subfloat[RGB image]{\includegraphics[width=\pipewidth]{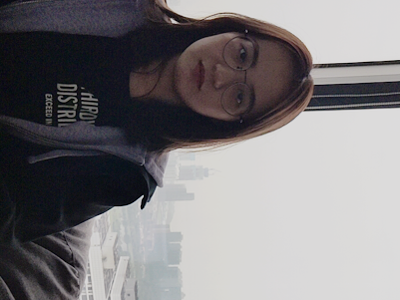}}\hspace{\inter}
    \subfloat[ToF amplitude]{\includegraphics[width=\pipewidth]{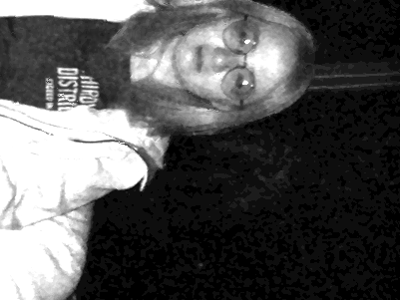}}\hspace{\inter}
    \subfloat[ToF depth]{\includegraphics[width=\pipewidth]{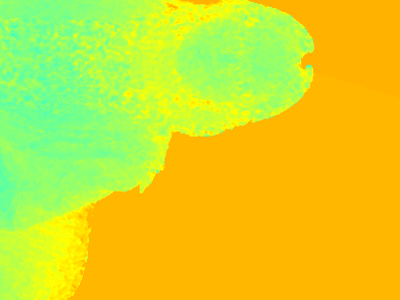}}\hspace{\inter}
    \subfloat[Results of DEAR]{\includegraphics[width=\pipewidth]{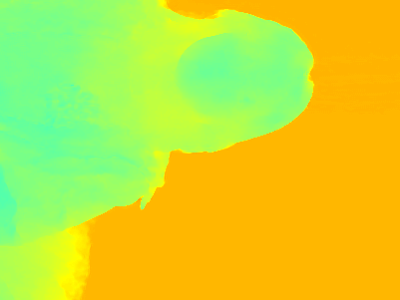}}\hspace{\inter}
    \subfloat[ToF depth + RGB]{\includegraphics[width=\pipewidth]{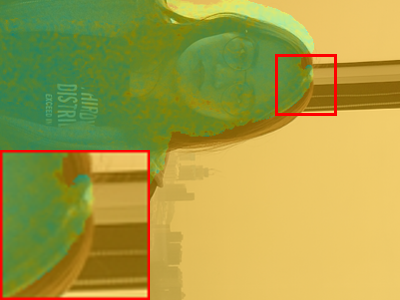}}\hspace{\inter}
    \subfloat[DEAR + RGB]{\includegraphics[width=\pipewidth]{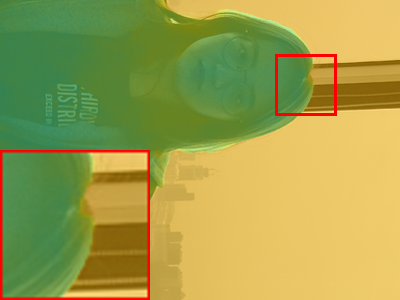}} \\[-2ex]
 \caption{Visual results of our deep end-to-end alignment and refinement framework. In the first three rows we show the results on synthetic data, while last three rows for real data taken by weakly calibrated ToF RGB-D camera modules.}
 \label{fig:supp_pipeline}
\end{figure*}

\subsection{More Visual Results on Flow Refinement}
More visual results of our optical flow refinement module on the ToF-FlyingThings3D dataset are presented in Figure\,\ref{fig:supp_flow_refine}.
We can see that, the quality of the optical flow is substantially improved by our flow refinement module.

\subsection{More Results on ToF-KPN}
We hereby show more depth image refinement results of ToF-KPN. Specifically, we apply it onto our ToF-FlyingThings3D dataset where each data instance are already aligned. 
Some of the results are shown in Figure\,\ref{fig:supp_depth_refine}.
It can be seen that our refinement results are very close to the ground-truth depth images.

We further demonstrate the MPI reduction of our ToF-KPN. 
Specifically, we present more comparisons to \textsc{DeepToF} \cite{marco2017deeptof} and the method of Su~{\it et al.}\;\cite{su2018deep} in Figure\,\ref{fig:supp_exp_tof}.
We follow the identical settings as in Section\,{5.3} of the paper, and plot the depth values along scan-lines of four different scenes.
We clearly see that, our ToF-KPN has greatly suppressed the MPI effects (compare to the original ToF depth images) while provides very high depth accuracies (compared to \textsc{DeepToF} \cite{marco2017deeptof} and Su~{\it et al.} \cite{su2018deep}).
\subsection{More Visual Results of DEAR}
More results of our DEAR framework are shown in Figure\,\ref{fig:supp_pipeline}, following the same settings of Section\,5.4 of the paper.
In Figure\,\ref{fig:supp_pipeline}, the first three rows show the results on the synthetic data while
the rest show results of our real data. 
It can be seen that, our DEAR framework provides visually pleasant depth results, which are not only well-aligned with the corresponding RGB images but also largely refined compared to the original ToF depth images.

{\small
\bibliographystyle{ieee_fullname}
\bibliography{ref}
}

\end{document}